%% file: main.tex
\documentclass{article}

\usepackage{palatino}
\usepackage{amssymb}
\usepackage{amsmath}
\usepackage{amsthm}
\usepackage[a4paper]{geometry}
\usepackage[numbers, compress]{natbib}
\usepackage[colorlinks,linktoc=true,
             citecolor= Cerulean!85!green!90!black,
             linkcolor=YellowOrange,
             urlcolor=RubineRed!85!black,
             ]{hyperref}
\usepackage{graphicx}

\usepackage{multicol}
\usepackage{multirow}
\usepackage{wrapfig}
\usepackage[utf8]{inputenc} 
\usepackage[T1]{fontenc}    
\usepackage{url}            
\usepackage{booktabs}       
\usepackage{amsfonts}       
\usepackage{nicefrac}       
\usepackage{microtype}      
\usepackage[table, dvipsnames]{xcolor}     
\usepackage{lmodern}           
\usepackage{caption}
\usepackage{authblk}

\usepackage{longtable}
\usepackage{verbatim}
\usepackage{array}
\usepackage{enumitem}

\usepackage{tcolorbox}
\usepackage{listings}
\usepackage{arydshln}
\newcommand{\x}{\mathbf{x}}
\newcommand{\y}{\mathbf{y}}
\newcommand{\X}{\mathbf{X}}

\newcommand{\Z}{\mathbf{Z}}
\newcommand{\C}{\mathbf{C}}

\newcommand{\bu}{\mathbf{u}}

\definecolor{darkgreen}{rgb}{0.0, 0.5, 0.5}
\definecolor{shadecolor}{rgb}{0.9, 0.9, 0.9}
\definecolor{listinggray}{gray}{0.9}
\definecolor{keywords}{rgb}{0.0, 0.0, 0.5}
\definecolor{comments}{rgb}{0.0, 0.5, 0.0}

\setlist[itemize]{leftmargin=*,topsep=0pt,itemsep=0pt,parsep=0pt}
\setlist[enumerate]{leftmargin=*,topsep=0pt,itemsep=0pt,parsep=0pt}

\let\titleold\title
\renewcommand{\title}[1]{\titleold{#1}\newcommand{\thetitle}{#1}}
\def\maketitlesupplementary
   {
   \newpage
   \begingroup 
       \centering
       \Large
       \textbf{\thetitle}\\\vspace{0.5em}
       Supplementary Material\\\vspace{1.0em}
   \endgroup 
   }

\title{GenoTEX: An LLM Agent Benchmark for Automated Gene Expression Data Analysis}

\author[1]{Haoyang Liu}
\author[1]{Shuyu Chen}
\author[2]{Ye Zhang}
\author[1]{Haohan Wang}
\affil[1]{University of Illinois Urbana-Champaign}
\affil[2]{Novartis Institutes for BioMedical Research}
\affil[1]{\texttt{\{hl57, shuyu5, haohanw\}@illinois.edu}}
\affil[2]{\texttt{ye-4.zhang@novartis.com}}

\date{}

\begin{document}

\maketitle

\begin{abstract}
\input{secs/abstract}
\end{abstract}

\section{Introduction}
\label{sec: intro}
\input{secs/intro}

\section{Related work}
\label{sec: related}

\input{secs/related}

\section{Benchmark}
\label{sec: Benchmark_generation}
\input{secs/benchmark}

\section{Method}
\label{sec: method}
\input{secs/method}

\section{Experiment}
\label{sec: Experiment}
\input{secs/experiment}


\section{Conclusion}
\label{sec: con}
\input{secs/con}

\section*{Acknowledgments and Disclosure of Funding}
\label{sec: ack}
\input{secs/ack}

\newpage 

\bibliography{ref}
\bibliographystyle{abbrvnat}

\newpage
\appendix
\onecolumn
\label{supplementary materials}
\input{secs/supplementary_materials}

\end{document}

%% file: secs/abstract.tex
Recent advancements in machine learning have significantly improved the identification of disease-associated genes from gene expression datasets. However, these processes often require extensive expertise and manual effort, limiting their scalability. Large Language Model (LLM)-based agents have shown promise in automating these tasks due to their increasing problem-solving abilities. To support the evaluation and development of such methods, we introduce GenoTEX, a benchmark dataset for the automated analysis of gene expression data. GenoTEX provides analysis code and results for solving a wide range of gene-trait association problems, encompassing dataset selection, preprocessing, and statistical analysis, in a pipeline that follows computational genomics standards. The benchmark includes expert-curated annotations from bioinformaticians to ensure accuracy and reliability. To provide baselines for these tasks, we present GenoAgent, a team of LLM-based agents that adopt a multi-step programming workflow with flexible self-correction, to collaboratively analyze gene expression datasets. 
Our experiments demonstrate the potential of LLM-based methods in analyzing genomic data, while error analysis highlights the challenges and areas for future improvement. We propose GenoTEX as a promising resource for benchmarking and enhancing automated methods for gene expression data analysis. The benchmark is available at \url{https://github.com/Liu-Hy/GenoTEX}.

%% file: secs/intro.tex
In biomedical research, gene expression analysis is crucial for understanding biological mechanisms and advancing clinical applications such as disease marker identification and personalized medicine. Advances in next-generation sequencing and other technologies have led to a surge in the volume of transcriptomic data. 
Genomics research is expected to produce between 2 and 40 exabytes of data in the next decade \cite{genomegov2024}, 
greatly facilitating research and discoveries in genomics.

Among the various approaches in computational biology, differential gene expression analysis identifies genes whose expression levels vary between biological conditions \cite{Love2014}. Our work focuses on gene-trait association (GTA) problems, which specifically examine how genes relate to traits (such as diseases) while accounting for additional biological factors or conditions \cite{GTEx2020, Mancuso2017}. These GTA problems enable important clinical applications \cite{Visscher2017}, where understanding gene expression patterns across diverse patient conditions can reveal context-dependent biomarkers and treatment targets \cite{Marigorta2018}.

Despite the scientific value of gene expression data analysis, these tasks are often repetitive, labor-intensive, and prone to errors \cite{RedR2023}. The rapid increase in transcriptomic data and potentially inefficient workflows lead to considerable financial burden \cite{Mordor2023}. The genetics research industry incurs an annual expense of around \$848.3 million on manual data analysis tasks \cite{researchmarkets2024}, with costs expected to increase at a compound annual growth rate of 12\% \cite{researchmarkets2024} to 16\% \cite{databridge2024} by 2030. Bioinformaticians spend significant effort on these repetitive tasks, valued at around \$29 per hour \cite{payscale_bioinfo}. This high volume of routine tasks greatly impacts job satisfaction among bioinformatics professionals, as surveys show that data scientists, including bioinformaticians, prefer engaging in advanced analytical tasks rather than routine data processing \cite{Alexw}. Currently, up to 45\% of their work hours are spent on tasks that could be automated \cite{Alexw}. These financial and workforce challenges highlight the urgent need for more efficient and cost-effective data analysis solutions in genetics research \cite{Kev2023}.

Meanwhile, the increasing abilities of Large Language Models (LLMs) \cite{openai2024gpt4o} have enabled them to act as agents to perform various data analysis tasks \cite{ma2023insightpilot, Arasteh2024}, and relevant benchmarks have been proposed \cite{stuhler2023benchmarking, eldeeb2024automlbench}. However, these studies have mostly either focused on solving specific problems in data analysis such as missing data imputation \cite{ding2024data, chen2023seed}, or end-to-end analysis on well-structured datasets with the goal of optimizing predictive accuracy \cite{grosnit2024large, guo2024dsagent, hong2024data}. In contrast, analysis on real-world gene expression data involves complex procedures for integrating data across \emph{multiple large, semi-structured files}, with the goal of finding significant variables from the high-dimensional, sparse data. This not only requires the relevant biological and statistical knowledge, but also the flexible planning, troubleshooting, and domain knowledge inference abilities typically performed by human bioinformaticians, posing higher demands to agentic methods.

\begin{figure*}
    \centering
    \includegraphics[width=0.9\textwidth]{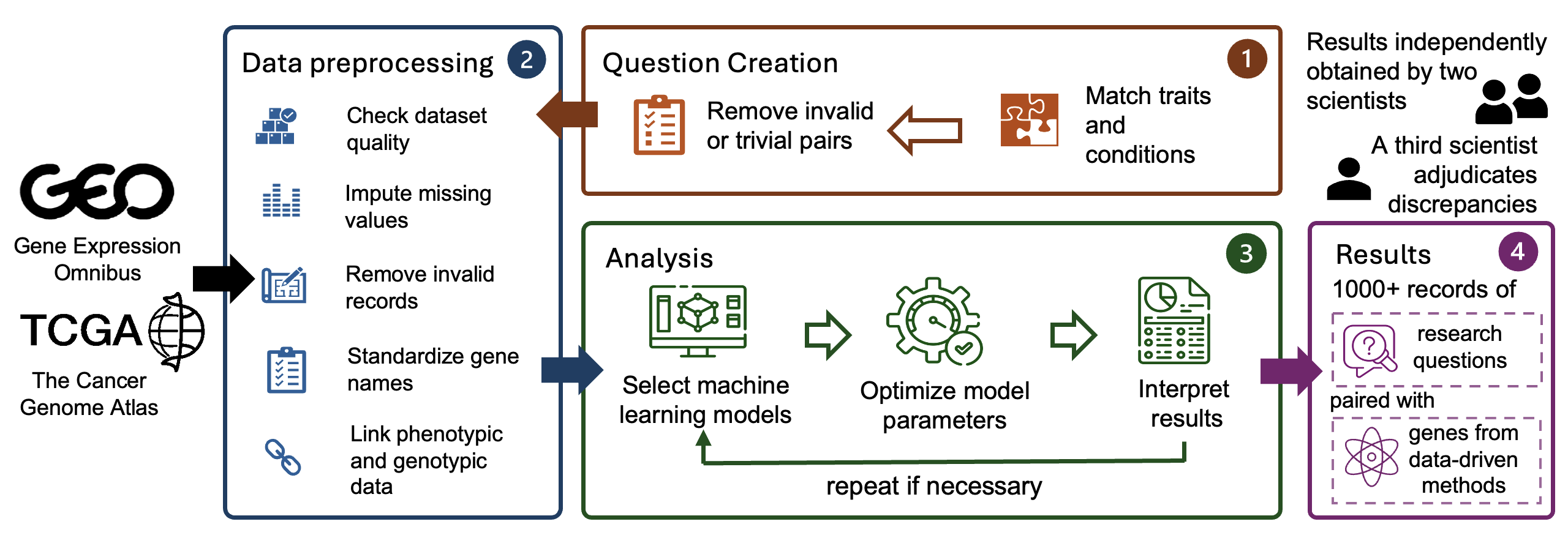}
    \caption{\footnotesize The overview of the GenoTEX benchmark curation, illustrating the standardized pipeline for analyzing gene expression datasets and the steps involved in creating the benchmark dataset.
    }
    \label{fig:main}
\end{figure*}

To facilitate the evaluation and development of such methods, we propose the \textbf{Geno}mics Data Au\textbf{t}omatic \textbf{Ex}ploration Benchmark (\textbf{GenoTEX}), a benchmark dataset for the automated analysis of gene expression datasets to solve GTA problems while considering the influence of other biological factors. Following the standards of computational genomics, and inspired by the behaviors of skilled human bioinformaticians, we unified the process of analyzing various gene expression datasets for GTA analysis into a standardized pipeline, with detailed procedures documented in a guidelines file (Appendix \ref{sec: guidelines}). We then trained and organized a group of bioinformaticians to manually perform the data analysis according to these guidelines, creating a benchmark dataset consisting of input data, annotated code, and intermediate and final analysis results. Fig.~\ref{fig:main} illustrates our benchmark curation pipeline. Based on this benchmark, we propose three tasks, namely, dataset selection, data preprocessing, and statistical analysis, with corresponding metrics for evaluating different aspects of the analysis process.
Furthermore, to provide baselines for these tasks, we propose GenoAgent, a team of LLM-based agents that simulate the behavior of bioinformaticians in gene data analysis. GenoAgent adopts a multi-agent collaboration pattern that facilitates task decomposition and code correction, to perform the end-to-end data analysis for solving GTA problems. Each programming agent is instructed with detailed guidelines to perform tasks in a Jupyter Notebook-style multi-step workflow.

Our evaluation suggests that GenoAgent is able to automate the process of gene expression data analysis with good overall accuracy, affirming the promise of integrating LLMs into genomics research. However, the observed significant gap from human expert performance and the reliance on advanced LLMs highlight the need for future research.

In summary, our contributions are as follows:
\begin{itemize}
    \item We propose a benchmark dataset, GenoTEX, that presents the analysis pipeline for a rich set of GTA problems, with documented code and output. 
    We believe it will serve as a useful resource for the evaluation and development of advanced methods for automated gene expression data analysis. 
    \item We define three challenging tasks: dataset selection, data preprocessing, and statistical analysis, along with corresponding metrics, to support more systematic evaluation. 
    \item We propose a baseline method, GenoAgent, a team of LLM-based agents to collaboratively explore gene expression datasets. Our evaluation demonstrates the promise of LLM-based approaches in genomics data analysis, and error analysis reveals areas for future improvement.
\end{itemize}

%% file: secs/related.tex
\paragraph{LLMs for collaborative problem-solving} Recent advancements in Large Language Models (LLMs) have demonstrated their potential to approach human-level intelligence in various domains \cite{wang2023survey, openai2023gpt4, touvron2023llama, touvron2023llama2}. Researchers have developed multiple strategies to enhance LLMs' problem-solving capabilities. These include goal decomposition approaches \cite{wei2022chain, zheng2023synapse, feng2023towards, ning2023skeleton}, which break complex problems into manageable sub-problems; structured reasoning methods using tree and graph representations \cite{yao2023tree, hao2023reasoning, besta2023graph}; and self-improvement techniques such as consistency checking \cite{wang2022self} and iterative refinement \cite{xi2023self, madaan2023self, wang2023recmind, chen2023teaching}. Integration with external tools has further expanded LLMs' functionality \cite{liu2023llmp, zhao2023verifyandedit, qin2023toolllm}.

Multi-agent collaboration presents a particularly promising direction for enhancing problem-solving capacity \cite{wang2023unleashing, talebirad2023multiagent, du2023improving, wang2023adapting}. Role-playing frameworks, where agents assume distinct areas of expertise, have proven effective in complex tasks \cite{yang2023autogpt, dong2023selfcollaboration}. For instance, some works \cite{hong2023metagpt, qian2023communicative} orchestrates collaboration among specialized agent roles, demonstrating the effectiveness of role-based collaboration in software development contexts. Recent work has focused on improving collaborative performance through task management and feedback mechanisms \cite{huang2023large, xu2023reasoning, gou2023critic, yin2023exchange}.

\paragraph{LLMs for scientific discovery} 
Researchers have increasingly integrated LLMs into scientific discovery workflows across multiple domains. These applications span chemistry \cite{bran2023chemcrow, guo2023can}, biotechnology \cite{liu2024team, madani2023large}, and medicine \cite{singhal2023large, yang2023large}. Most of them have relied on training or fine-tuning LLMs with domain-specific data to achieve specialized capabilities.

Our work leverages state-of-the-art LLMs without additional domain-specific training. Instead, we develop structured prompting techniques and communication protocols that enable LLM-based agents to perform sophisticated planning, analysis, and coding tasks required for genomic data exploration. This approach demonstrates how general-purpose LLMs can be effectively applied to specialized scientific domains through careful system design rather than model customization.


%% file: secs/benchmark.tex
This section describes our GenoTEX benchmark. Specifically, we introduce our standardized pipeline for gene expression data analysis, the benchmark curation process with quality control, and the tasks and metrics defined for evaluation.

\subsection{Standardized pipeline of gene expression data analysis}
Our study aims to automate the analysis of gene expression data to address a class of important problems: \emph{What are the significant genes associated with a specific trait, considering the influence of some condition?} Here, a ``trait'' refers to a characteristic such as a disease (e.g., diabetes), and a ``condition'' refers to a factor like age, gender, or a co-existing trait (e.g., hypertension). This problem is scientifically important because the key genes linked to traits often vary based on the diverse physical conditions of patients. By incorporating these factors into our analysis, we gain a more comprehensive gene-trait knowledge profile, which mimics the real-life research scenario.

Evaluating the automatic exploration of this kind of problems is complex due to its nature. The combination of different traits and conditions leads to a multitude of GTA problem scenarios, many of which remain understudied in biomedical literature. Moreover, while data-driven approaches provide valuable insights, they must ultimately be combined with interventional biological experiments or clinical trials to confirm the significance of identified genes. This absence of a clear ``ground truth'' complicates the evaluation of our analysis results. Defining the exact insights that should be extracted from our data analysis is therefore complex.

To address these challenges, we designed the benchmark by collecting analysis outputs from skilled bioinformaticians following a standardized analysis pipeline that represents state-of-the-art analytics. This standardized approach enables systematic evaluation of the automated methods against established human expertise, aiming to facilitate not only the evaluation of current agent methods but also future more advanced methods.

In the following subsection, we introduce this pipeline in detail and provide the necessary background knowledge to understand its significance and application in our research.

\label{sec: pipeline}
\subsubsection{Data preprocessing}
Gene expression data preprocessing aims to extract, normalize, and integrate data across multiple large, semi-structured files, encompassing several main steps such as dataset filtering and selection, gene data preprocessing, trait data extraction, and data linking. In our paper, we use ``dataset'' to refer to a cohort dataset, which is the collection of samples with associated genetic and clinical information from a biomedical study. We also use ``benchmark dataset'' to refer to the collection of all such datasets and their analysis results. Their distinction should be clear from the context. Below we introduce our preprocessing steps in more details.

\paragraph{Dataset filtering and selection} 
This step aims to filter out irrelevant datasets and select the best dataset (or pair of datasets) to analyze for each GTA problem. The selection of datasets involves three steps: 
\begin{enumerate}
\item \textbf{Initial filtering } We start with a list of potentially useful datasets, and determine the relevance of each dataset to the problem by reading the metadata. This involves verifying the availability of gene expression data (as opposed to miRNA or methylation data) and the traits of interest. 
\item \textbf{Quality verification } In case there are abnormalities in the dataset that were not handled successfully in the preprocessing step, we discard the dataset to ensure quality. 
\item \textbf{Dataset selection } As gene expression data are often high-dimensional and scarce, the analysis can be bottlenecked by sample size. Therefore, if multiple preprocessed datasets are available for statistical analysis about a trait, we select the one with the largest sample size. If the analysis requires integrating datasets about two traits, we sort the possible pairs of datasets for both traits by the product of their sample sizes, and select the pair with the largest product.
\end{enumerate}
\paragraph{Gene expression data preprocessing} This step aims to prepare a gene expression data table, where each attribute represents the expression level of a specific gene within samples. The procedure varies based on the measurement technique. For microarray data, we start with raw datasets identified by probe IDs, which are DNA sequences complementary to the target RNA sequences used to measure gene expression. For RNA-seq data, we handle sequence reads that require alignment to a reference genome.
In both cases, we map initial identifiers to gene symbols using platform-specific gene annotation data. We then normalize and deduplicate these gene symbols by querying a local gene database, to prevent inaccuracies arising from different gene naming conventions. This process requires flexible planning and proficient use of bioinformatics tools to ensure accuracy and consistency.
\paragraph{Trait data extraction} This step aims to prepare a data table recording the phenotypic trait and demographic information (i.e. age, gender) of samples. Clinical information typically appears in variably positioned rows or columns of raw data tables, with attribute names and values in dataset-specific free-text forms. To get standardized trait data, we need to identify the attributes containing the variable information of interest, design conversion rules, and write functions to encode the attributes into binary or numerical variables. Often this information is indirectly given, requiring us to infer it based on an understanding of the data measurement and collection process described in the metadata, combined with necessary domain knowledge about related jargon or acronyms. Some examples of this step are shown in Appendix \ref{sec: examples}.
\paragraph{Data linking}
This step aims to link the preprocessed gene data with the extracted trait data based on sample IDs. This integration creates a data table containing both genetic and clinical features for the same samples, ready for association studies to identify significant genes. \\

\begin{figure*}
    \centering
    \includegraphics[width=0.95\textwidth]{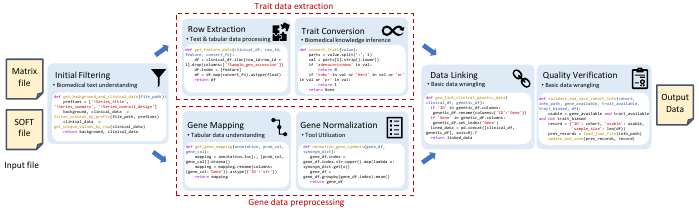}
    \caption{High-level schematic of our GEO data preprocessing pipeline, with example code of core components that omits techinical details. }
    \label{fig: preprocess_pipeline}
\end{figure*}

The preprocessing also involves common operations such as missing value imputation and column matching, some of which are substeps of the main steps. Our guidelines file in Appendix~\ref{sec: guidelines} provides additional details. Fig. \ref{fig: preprocess_pipeline} illustrates the pipeline for preprocessing a series dataset from the GEO database. 

\subsubsection{Statistical analysis}
After preprocessing, one can perform basic regression analysis to identify genes associated with the disease (or trait) \cite{ghosh2005classification,wu2009genome}. 
Despite the popularity of deep learning methods and their variants, the genetic community typically prefers linear methods due to their interpretability \cite{cook2017guidance} and lower sample-size requirements \cite{angermueller2016deep}. 
Lasso \cite{tibshirani1996regression} is often chosen as the model due to its ability to identify a sparse set of genes. 
In addition to directly applying regression models, some additional analytical steps are often taken. We provide mathematical details of these steps in Appendix \ref{sec: statistics}.
\paragraph{Confounding factor correction}
To ensure reliable GTA analysis, the pipeline incorporates steps to correct potential confounding factors \cite{bruning2016confounding}, including batch effects \cite{leek2010tackling}. One important type of confounding arises when the distribution of gene expressions varies across subgroups within the data due to sample population heterogeneity rather than the disease itself \cite{yu2006unified}. This variation introduces significant bias, as spurious gene-disease associations may be detected due to technical or population differences rather than genuine biological relationships \cite{wang2022trade}. We directly use a Python software package developed to handle different scenarios of confounding factors \cite{wang2022trade}.

\paragraph{Incorporating conditions in regression}
The analysis is enhanced by including additional covariates in the regression model, such as patient demographics and comorbidities \cite{kyalwazi2023race}. 
This step allows for identifying gene expression patterns that are not only associated with the disease status but also modulated by these conditions. 
To implement this, we expand the feature set to incorporate these conditions and build regression on this expanded feature set.

\subsection{Benchmark creation}
\label{sec: benchmark_creation}
\begin{wrapfigure}{r}{0.52\textwidth}
\vspace{-5.2mm}
\small
\centering
\captionof{table}{\small \textbf{Descriptive statistics of our GenoTex benchmark.}}
\begin{tabular}{p{0.65\linewidth}r}
\toprule
\multicolumn{2}{c}{\textbf{GTA Problems}} \\
\midrule
Total problems & 1384 \\
Unconditional problems & 132 \\
Conditional problems & 1252 \\
\midrule
\multicolumn{2}{c}{\textbf{Input dataset}} \\
\midrule
Total size & 41.5 GB \\
Datasets & 911 \\
Samples per dataset & 167 $\pm$ 121\\
Total samples & 152,415 \\
\midrule
\multicolumn{2}{c}{\textbf{Manual Analysis and Results}} \\
\midrule
Relevant datasets & 467 \\
Datasets successfully preprocessed & 448 \\
Lines of code for analyzing per dataset & 261 $\pm$ 108 \\
Total lines of code for analysis & 237,907 \\
Normalized gene features per dataset & 18530 $\pm$ 6714 \\
Significant genes identified per problem & 359 $\pm$ 265 \\
\bottomrule
\end{tabular}
\vspace{-4.8mm}
\end{wrapfigure}
Given the complexity of the above-mentioned data analysis pipeline, the systematic evaluation of automated analysis methods necessitates a high-quality, multi-task benchmark. This subsection describes our process for creating GenoTEX, including problem design, genomic data acquisition, and our rigorous manual curation workflow.

\paragraph{GTA problem design}
We began by curating a scientifically relevant list of human traits spanning from major public health concerns to representative rare diseases, along with physical characteristics like height and bone density. A computational biologist compiled 132 traits across 9 main categories, including cardiovascular diseases and neurological disorders. This yielded our first set of problems: \emph{What are the significant genes related to the trait?} (unconditional GTA problem).

We then created trait-condition pairs by matching each trait with a condition—either another trait from our list or demographic attributes like age and gender. From the 17,556 possible trait-condition pairs, we selected the most scientifically valuable ones to frame our second problem set: \emph{What are the significant genes related to the trait when considering the influence of the condition?} (conditional GTA problem).

Our selection of trait-condition pairs involved both domain expertise and data-driven approaches. We applied manually designed rules determining which pairs to include or exclude based on trait categories (detailed in Appendix \ref{sec: criteria}). For undecided pairs, we measured trait-condition association by calculating Jaccard similarity between gene sets related to each trait and condition, using data from the Open Targets Platform \cite{opentargets}. We selected pairs with Jaccard similarity exceeding 0.1, as these likely share underlying genetic mechanisms, offering valuable insights into complex trait-condition interactions. This process yielded 1,252 conditional GTA problems of significant scientific interest. Combined with our 132 unconditional problems, these constitute our benchmark's complete problem set.

\paragraph{Input dataset}
To address our research problems, we sourced cohort datasets containing gene expression and clinical data from leading public repositories: (1) The Gene Expression Omnibus (GEO) \cite{clough2016gene}, the largest available gene expression database; (2) The Cancer Genome Atlas (TCGA) \cite{tomczak2015cancer}, the leading cancer-focused gene expression database, accessed via the UCSC Xena platform \cite{ucsc_xena}. Additionally, we acquired essential domain knowledge from the Open Targets Platform \cite{opentargets} for gene-trait associations and the NCBI Genes database \cite{NCBIGene} for gene synonym mapping. Appendix \ref{sec: data_acquisition} provides detailed information about our data acquisition process.

\paragraph{Manual analysis}
Our team of 4 researchers designed the problem list and acquired relevant input data from public sources. During the pilot stage, a computational biologist and a doctoral student collaboratively developed example code for solving GTA problems related to two traits. By extracting common patterns from these examples, they developed an initial version of the analysis guidelines for the entire benchmark. Then they iteratively refined the guidelines based on their experience analyzing 193 more problems, establishing a working framework for benchmark curation.

For comprehensive benchmark curation, we assembled and trained a team of 9 bioinformaticians to analyze the complete set of problems. Following the guidelines and example code, these bioinformaticians wrote programs to solve the problems assigned to them by analyzing relevant gene expression datasets, and submitted their analysis code and results weekly over a period of 20 weeks. We implemented a continuous feedback system where bioinformaticians documented analysis challenges and edge cases they encountered, discussed potential solutions through online channels, and participated in weekly team meetings to share their insights. Based on these discussions, our core team of four researchers refined the guidelines to address corner cases and dataset format idiosyncrasies, with the bioinformaticians subsequently updating their previous analyses to reflect guideline improvements. Each trait was independently analyzed by two bioinformaticians, with an experienced researcher reviewing both versions to select the superior analysis and make additional refinements.

To evaluate annotation consistency, we measured Inter-Annotator Agreement (IAA) between the two annotation versions. Results demonstrate high quality, with an F$_1$ score of 94.73\% for dataset filtering. For more evaluation of the benchmark quality, we take the analysis results in the final benchmark as the reference set, and computed the average performance of the two annotation versions using the metrics defined in Section \ref{sec: tasks}. This also serves as a strong baseline of human expert performance, with results presented in Table~\ref{tab: end2end_main}. The set of identified significant genes is sensitive to specific choices made during cohort-specific feature encoding, where multiple reasonable approaches often exist. Despite this inherent ambiguity in gene selection, the AUROC score of 0.89 demonstrates high consistency among annotators, validating the reliability of our benchmark.

\subsection{Tasks and metrics}
\label{sec: tasks}
\paragraph{Dataset selection and filtering}
We evaluate Dataset Filtering (DF) and Dataset Selection (DS) as distinct tasks. DF is a binary classification task predicting whether a dataset is usable, and we use F$_1$ as the primary metric. For DS, we use accuracy to measure the percentage of problems where the method selects the same dataset (or pair of datasets) as specified in our benchmark.
\paragraph{Preprocessing} Gene expression data preprocessing involves complex decisions that affect both attribute names and sample IDs. We evaluate preprocessing performance using three complementary metrics: (i) Attribute Jaccard (AJ) is the Jaccard similarity between sets of attributes of two datasets. It evaluates how well the method extracts attributes from the dataset by encoding clinical features and normalizing gene symbols. (ii) Sample Jaccard (SJ) is the Jaccard similarity between sets of sample IDs of two datasets. It measures how well the method integrates features from the same samples and handles missing values. Based on these metrics, we define (iii) \textbf{Composite Similarity Correlation (CSC)} as the product of the Attribute Jaccard, Sample Jaccard, and the average Pearson correlation of common feature vectors (shared rows and columns) between datasets. This metric captures both structural and numerical similarity of datasets, so we consider it as the primary metric for evaluating preprocessing quality. We compute these scores per dataset and average across all datasets in our benchmark.
\paragraph{Statistical analysis}
Statistical analysis in our benchmark identifies trait-related significant genes, which we evaluate using multiple complementary metrics. We frame GTA analysis as a binary classification task, where each gene is classified as either significant or non-significant to a problem. We assess the classification performance using Precision, Recall, and F$_1$ scores. To address the challenge of imbalanced gene sets, we employ the Area Under the Receiver Operating Characteristic curve (AUROC) \cite{hanley1982meaning}, which provides a threshold-independent measure of discrimination ability. Additionally, we incorporate the Enrichment Score (ES) from Gene Set Enrichment Analysis (GSEA) \cite{subramanian2005gene}, a standard bioinformatics metric that captures the biological relevance of identified gene sets. We designate AUROC as our primary metric due to its robustness to class imbalance and widespread acceptance in the machine learning community. All metrics are computed per problem and averaged across all problems in our benchmark.

%% file: secs/method.tex
Recent studies have explored using LLM-based agents to tackle challenging problems \cite{huang2023large, yin2023exchange}, including various data analysis tasks \cite{ma2023insightpilot, Arasteh2024}. While these methods offer innovative approaches with distinct strengths, our preliminary experiments (in Appendix \ref{sec: attempts}) reveal that none successfully generate functional code capable of running end-to-end data analysis on our benchmark, after extensive attempts and given detailed instructions and function tools. This is not surprising given the intricate complexity of analyzing real-world gene expression data, which necessitates a more specialized approach. This section presents GenoAgent, our method for establishing a baseline for gene expression data analysis tasks.

\subsection{Motivation and agent design}
When a human expert writes programs for gene expression data analysis, they exhibit the following abilities: (i) \textbf{Context-aware planning.} 
They complete a task step by step, choosing the next action based on the overall goal and the results of previous steps; (ii) \textbf{Tool utilization.} They select and use library functions to assist with data preprocessing and statistical analysis; 
(iii) \textbf{Domain knowledge inference.} They observe the metadata of the dataset and intermediate processing results, using domain knowledge to infer the desired information from the data and use these observations to check whether their code works as expected; (iv) \textbf{Error correction.} they analyze the errors in program execution and correct them. 

We believe that integrating these components is essential for enabling agent systems to effectively tackle the complex task of gene expression data analysis. Thus, to propose reasonable baselines for our benchmark, inspired by human bioinformaticians' workflow in gene data analysis, we propose GenoAgent, a team of LLM-based agents with specialized roles mirroring those in a genomic data science team.
A \emph{Project Manager} coordinates the analysis process for solving each GTA problem, assigning tasks to agents with the standardized pipeline in our benchmark as instructions. Two programming agents, \emph{Data Engineer} and \emph{Statistician}, focus on the data preprocessing and statistical analysis tasks, respectively. 
When writing code to perform specific steps, they can access, adapt, and utilize function tools from a provided library file as needed. A \emph{Code Reviewer} agent helps the programming agents debug code and verify that their code follows the instructions. A \emph{Domain Expert} agent provides professional knowledge consultation to programming agents when required for data processing. 

\paragraph{Context-aware planning}
To enable effective planning within this complex analytical workflow, our agents maintain a comprehensive task context that records the text instruction, code, and execution output for each previous step. Before proceeding, agents observe the current task context and determine whether to perform the next step, skip it, or revert to a previous step if necessary. This approach provides the flexibility needed to navigate the intricate, multistep process of gene expression analysis, where errors can cascade catastrophically if not properly managed.

\subsection{Collaboration among LLM agents}
\label{sec:collaboration}
This subsection introduces the two main patterns of collaboration between agents.

\paragraph{Code review and iterative debugging}
This collaboration pattern involves interaction between the Code Reviewer and a programming agent (Statistician or Data Engineer). When code execution fails, the Code Reviewer evaluates the code based on its execution result, error trace, and compliance with the given instructions. The reviewer then decides to either approve the code or reject it with detailed feedback for revision, as illustrated in Figure~\ref{fig:reviewer} in the Appendix. Based on this feedback, the programming agent iteratively refines the code, extending the context with new versions until receiving approval or reaching the maximum debugging iterations. This mechanism not only facilitates troubleshooting but also ensures adherence to task instructions.

\paragraph{Domain-guided programming}
The second collaboration pattern involves a Data Engineer consulting a Domain Expert for data preprocessing tasks that require specialized knowledge. The Data Engineer sends questions to the Domain Expert, providing necessary context such as dataset metadata, summary information, or other intermediate processing results. The Domain Expert then provides answers in the form of executable code. This type of programming also undergoes a debugging process, but the execution results are sent back to the same Domain Expert rather than a separate reviewer. Some questions are complex enough that the Domain Expert may not provide a correct answer immediately, necessitating further refinement based on execution results.

%% file: secs/experiment.tex
This section describes our experiments to evaluate GenoAgent and other baseline methods on the GenoTEX benchmark. We conducted an end-to-end evaluation where methods process raw input data to complete the full analysis for solving GTA problems. Additionally, we assessed these methods on each individual task to identify specific strengths and weaknesses. All evaluations use the tasks and metrics defined in Section~\ref{sec: tasks}. We tested GenoAgent on a wide range of representative LLMs, as shown in Table~\ref{tab: end2end_main}.

\paragraph{Computation environment}  
All experiments were conducted on two instances of the RunPod GPU Cloud \cite{runpod}, each equipped with 16 vCPU cores and 250GB RAM. The execution of data analysis code generated by the agents required approximately 0.3GB CPU RAM for dataset processing, independent of the LLM used for code generation.

We employed different deployment strategies for LLM inference. For the 70B parameter models (\texttt{Llama 3.3-70B} and \texttt{DeepSeek R1-70B}), we used the Ollama \cite{Ollama} library to run inference locally on an H100 NVL GPU with 94GB memory, observing peak VRAM consumption of around 75.8GB. For all other LLMs, we utilized remote API services. Specifically, proprietary models like \texttt{OpenAI o1} and \texttt{Claude 3.5 Sonnet} were accessed through their respective official APIs. For the full 671B versions of \texttt{DeepSeek-V3} and \texttt{DeepSeek-R1}, we used the API service from Novita AI \cite{novitaai} rather than DeepSeek's official endpoints due to their lower latency.

\begin{table*}[ht!]
\centering
\caption{\textbf{End-to-end performance of GenoAgent on various LLMs; ablation results of GenoAgent with OpenAI o1 model; human expert performance, and simple data-free baselines.}
We report performance on the GTA problems in our benchmark, on the additional task of trait prediction, and runtime performance including success rate and efficiency metrics. Success rate (Succ.) refers to the percentage of problems for which the generated analysis code executes successfully. We use $0$ as the trait prediction scores when code execution fails. ``Tk.(K)'' reports average input/output tokens (in thousands) per problem, while ``Cost (\$)'' and ``Time (s)'' represent average API cost and runtime per problem.
} 
\label{tab: end2end_main}
\renewcommand{\arraystretch}{1.1}
\resizebox{\linewidth}{!}{
\setlength{\tabcolsep}{1.32mm}{
\begin{tabular}{lccccc@{\hspace{5.5mm}}cc@{\hspace{5.5mm}}cccc}
\toprule
 \multirow{2}{*}{Model} 
 & \multicolumn{5}{c@{\hspace{5.5mm}}}{GTA Analysis} 
 & \multicolumn{2}{c@{\hspace{5.5mm}}}{Trait Prediction} 
 & \multicolumn{4}{c}{Runtime Performance} \\
\cmidrule(l{3.5pt}r{5.5mm}){2-6} \cmidrule(r{5mm}){7-8} \cmidrule(r){9-12}

 & Prec.(\%) & Rec.(\%) & F$_1$(\%) & AUROC & GSEA & Acc.(\%) & F$_1$(\%) & Succ.(\%) & Tk.(K) & Cost(\$) & Time(s) \\
\midrule
\rowcolor[HTML]{E3F2FD}[8pt]
OpenAI o1 \cite{openai2024o1}              & \textbf{43.26} & \textbf{50.33} & \textbf{41.39} & \textbf{0.74} & \textbf{0.45} & \textbf{85.76} & 77.29 & 76.23 & 112.3/12.1 & 1.96 & 192.43 \\
OpenAI o3-mini \cite{openai2025o3mini}     & 34.32 & 37.90 & 31.04 & 0.70 & 0.34 & 77.01 & 71.19 & 58.16 & \textbf{108.9}/20.2 & 0.20 & 137.38 \\
GPT-4o \cite{openai2024gpt4o}              & 26.80 & 31.06 & 25.29 & 0.67 & 0.27 & 81.24 & 78.59 & 14.35 & 116.7/3.6 & 0.33 & \textbf{131.26} \\
Claude 3.5 Sonnet \cite{claude2024sonnet}  & 33.77 & 34.38 & 33.88 & 0.71 & 0.34 & 84.77 & 76.20 & \textbf{80.89} & 180.0/\textbf{2.7} & 0.58 & 134.73 \\
Claude 3.5 Haiku \cite{claude2024haiku}    & 19.54 & 17.32 & 14.79 & 0.63 & 0.16 & 79.64 & 78.86 & 19.71 & 204.3/3.1 & 0.18 & 188.52 \\
DeepSeek R1 \cite{deepseek2025r1}          & 38.65 & 30.38 & 30.73 & 0.70 & 0.31 & 85.10 & \textbf{78.96} & 9.04 & 120.9/35.4 & 0.63 & 923.09 \\
DeepSeek V3 \cite{deepseek2024v3}          & 39.80 & 23.29 & 23.09 & 0.67 & 0.23 & 83.02 & 75.91 & 4.83 & 115.8/3.8 & 0.11 & 245.19 \\
Gemini 2.0 Flash \cite{gemini2024flash}    & 15.37 & 27.56 & 15.10 & 0.65 & 0.19 & 74.21 & 71.28 & 6.85 & 304.6/16.8 & \textbf{0.06} & 186.15 \\
Gemini 1.5 Pro \cite{gemini2024pro}        & 12.53 & 15.67 & 11.52 & 0.62 & 0.13 & 78.46 & 76.45 & 13.62 & 163.1/\textbf{2.7} & 0.21 & 305.47 \\
Llama 3.3-70B \cite{llama370B}             &  3.13 &  3.94 &  3.02 & 0.52 & 0.04 & 71.25 & 65.54 &  2.61 & 189.4/5.8 & - & 319.37 \\
DeepSeek R1-70B \cite{deepseek2025r1}      &  3.76 &  4.77 &  3.62 & 0.53 & 0.04 & 68.36 & 62.81 &  3.17 & 158.5/14.3 & - & 721.86 \\
\addlinespace[1.7pt]
\hdashline[1.5pt/2pt]
\addlinespace[1.7pt]
No planning                                & 38.49 & 45.02 & 37.11 & 0.71 & 0.41 & 78.59 & 71.01 & 69.48  & 89.5/9.8 & 1.65 & 146.25 \\
No Domain Expert                           & 34.99 & 41.40 & 33.76 & 0.69 & 0.37 & 75.61 & 67.63 & 70.35  & 97.8/10.5 & 1.84 & 177.04\\
Review Round=1                             & 40.18 & 47.22 & 38.96 & 0.74 & 0.42 & 80.89 & 72.91 & 73.19  & 99.4/9.9 & 1.73 & 158.79\\
No reviewer                                & 13.87 & 17.37 & 14.58 & 0.23 & 0.15 & 27.39 & 24.64 & 30.51  & 84.9/9.1 & 1.51 & 121.23 \\
\addlinespace[2pt]
\hdashline[4.2pt/3pt]  
\addlinespace[2pt]
Human expert                               & 73.30 & 85.41 & 70.51 & 0.89 & 0.74 & - &  - &  -  &  -  &  -   & \\
\addlinespace[1.7pt]
\hdashline[1.5pt/2pt]
\addlinespace[1.7pt]
o1 zero-shot                               & 13.28 &  3.61 &  4.29 & 0.55 & 0.08 & - &  - &  -  & 0.1/0.6 & 0.04 & 10.45 \\
Random                                     &  0.72 &  0.90 &  0.75 & 0.49 & 0.00 & - &  - &  -  &  -  &  -   & 0.01 \\
\bottomrule
\end{tabular}
}
}
\end{table*}

\begin{table*}[ht!]
\centering
\caption{\textbf{Statistical analysis performance when using \emph{expert-preprocessed data} as input.} Evaluation metrics follow those in Table~\ref{tab: end2end_main}. Baselines include several representative agentic methods and LLMs, and fixed Python scripts implementing two standard regression models for gene expression analysis. ``BEC'' in Row 2 refers to batch effect correction.} 
\label{tab: regression}
\renewcommand{\arraystretch}{1.1}
\resizebox{\linewidth}{!}{
\setlength{\tabcolsep}{1.32mm}{
\begin{tabular}{lccccc@{\hspace{5.5mm}}cc@{\hspace{5.5mm}}cccc}
\toprule
 \multirow{2}{*}{Model} 
 & \multicolumn{5}{c@{\hspace{5.5mm}}}{GTA Analysis} 
 & \multicolumn{2}{c@{\hspace{5.5mm}}}{Trait Prediction} 
 & \multicolumn{4}{c}{Runtime Performance} \\
\cmidrule(l{3.5pt}r{5.5mm}){2-6} \cmidrule(r{5mm}){7-8} \cmidrule(r){9-12}
 & Prec.(\%) & Rec.(\%) & F$_1$(\%) & AUROC & GSEA & Acc.(\%) & F$_1$(\%) & Succ.(\%) & Tk.(K) & Cost(\$) & Time(s) \\
\midrule
\rowcolor[HTML]{E3F2FD}[8pt]
GenoAgent (Ours)                      & \textbf{94.81} & \textbf{92.27} & \textbf{93.83} & \textbf{0.97} & \textbf{0.95} & \textbf{91.95} & \textbf{90.22} & \textbf{100.00} & 15.1/3.6 & 0.44 & 58.41 \\
GenoAgent (No BEC)                    & 68.39 & 73.26 & 67.65 & 0.86 & 0.71 & 87.34 & 73.18 & \textbf{100.00} & 14.6/3.5 & 0.42 & 56.16 \\
Data Interpreter \cite{hong2024data}  & 90.73 & 89.07 & 89.52 & 0.95 & 0.92 & 87.84 & 86.03 & 98.45 & 26.0/6.1 & 0.74 & 167.54 \\
MetaGPT \cite{hong2023metagpt}        & 88.97 & 87.54 & 87.59 & 0.95 & 0.91 & 85.25 & 84.16 & 95.87 & 21.8/5.2  & 0.65 & 142.25 \\
CodeAct \cite{wang2024executable}     & 83.27 & 80.79 & 81.23 & 0.92 & 0.87 & 78.91 & 77.45 & 85.76 & 19.3/4.7 & 0.60 & 94.38 \\
OpenAI o1 \cite{openai2024o1}         & 79.95 & 78.48 & 78.66 & 0.91 & 0.84 & 75.98 & 74.67 & 82.13 & 9.1/3.2 & \textbf{0.28} & 34.06 \\
Llama 3.3-70B \cite{llama3}           & 56.08 & 53.29 & 54.01 & 0.82 & 0.56 & 72.62 & 70.95 & 68.25 & \textbf{7.4}/\textbf{0.9} & - & \textbf{21.72} \\
\addlinespace[1.7pt]
\hdashline[4.2pt/3pt]
\addlinespace[1.7pt]
Lasso \cite{tibshirani1996regression} & 32.40 & 13.02 & 14.37 & 0.53 & 0.12 & 84.10 & 64.27 & 99.31 & - & - & 10.40 \\
LMM \cite{henderson1950estimation}    & 3.70 & 3.73 & 3.71 & 0.53 & 0.02 & 75.06 & 59.72 & 99.31 & - & - & 18.52 \\
\bottomrule
\end{tabular}
}
}
\end{table*}

\subsection{Results}
\paragraph{End-to-end performance}
We evaluated the end-to-end data analysis capabilities of GenoAgent by measuring how effectively it identifies significant genes from raw input data. As shown in Table~\ref{tab: end2end_main}, GenoAgent achieved an AUROC score of $0.74$. While this represents promising performance given the task complexity, a significant gap remains compared to human expert performance ($0.89$ AUROC). The results also demonstrate a strong dependence on the underlying LLM capabilities, indicating substantial room for improvement.

Our ablation studies confirmed the benefits of our context-aware planning mechanism compared to executing all steps in predefined order. The results also highlighted the importance of collaborative approaches involving Code Reviewer and Domain Expert agents, as well as the value of multiple review rounds.

For comparison, we implemented a simple baseline where an \texttt{OpenAI o1} model was directly prompted to identify significant genes for each problem without access to any datasets or additional information. This approach yielded poor performance (0.55 AUROC), underscoring the difficulty of this task. Beyond GTA analysis, we also measured the agents' ability to predict binary traits using gene expression features and other covariates from their selected datasets, with strong results confirming the validity of the regression models they implemented.

\paragraph{Dataset filtering and selection} 
Table~\ref{tab: data_selection} presents the performance of dataset filtering and selection tasks. The agents demonstrated reasonable effectiveness in these tasks, likely because determining dataset relevance from metadata typically requires less complex inference than subsequent analysis steps. However, errors at this stage propagate through the pipeline and significantly impact overall performance.

\begin{table}[ht]
\centering
    \footnotesize
    \caption{\textbf{Performance of GenoAgent on dataset filtering and selection.}
    We report the F$_{1}$ score for Dataset Filtering (DF) and accuracy for Dataset Selection (DS).}
    \renewcommand{\arraystretch}{1.1}
    \begin{tabular}{lcc}
    \toprule
    Method & DF (\%) & DS (\%) \\
    \midrule
    \rowcolor[HTML]{E3F2FD}[6pt]
    GenoAgent (Ours) & \textbf{87.32} & \textbf{80.25} \\
    Rounds=1         & 85.29 & 76.04 \\
    No Reviewer      & 82.13 & 69.57 \\
    No Domain Expert & 84.28 & 78.63 \\
    \bottomrule
    \end{tabular}
    \label{tab: data_selection}
\end{table}

\paragraph{Dataset preprocessing} 
We evaluated the preprocessing performance of GenoAgent by comparing its output with that of human bioinformaticians in our benchmark. As shown in Table~\ref{tab: preprocess}, GenoAgent performed well in preprocessing gene expression data, achieving high CSC scores (80.63\% for genes). However, preprocessing trait data proved considerably more challenging, with a significantly lower CSC score of 32.28\%. This disparity stems from the complexity of clinical data extraction and the requirement for nuanced domain knowledge. Ablation results again demonstrated the importance of multi-agent collaboration and multiple review rounds for effective preprocessing.

\begin{table}[t]
    \centering
    \footnotesize
    \caption{\textbf{Performance of GenoAgent on the preprocessing tasks.} 
    Metrics reported for Linked, Gene, and Trait Data.}
    \begin{tabular}{llccc}
        \toprule
        Data Type & Method & AJ(\%) & SJ(\%) & CSC(\%) \\
        \midrule
        \rowcolor[HTML]{E3F2FD}
        \multirow{4}{*}{Linked Data} 
        & GenoAgent (Ours) & \textbf{89.82} & \textbf{86.98} & \textbf{79.71} \\
        & Round=1          & 87.04 & 82.15 & 74.43 \\
        & No Reviewer      & 35.18 & 35.06 & 32.73 \\
        & No Domain Expert & 78.54 & 75.93 & 70.01 \\
        \midrule
        \rowcolor[HTML]{E3F2FD}
        \multirow{4}{*}{Gene Data} 
        & GenoAgent (Ours) & \textbf{92.80} & \textbf{89.87} & \textbf{80.63} \\
        & Round=1          & 88.04 & 82.34 & 76.11 \\
        & No Reviewer      & 36.01 & 35.70 & 33.62 \\
        & No Domain Expert & 80.79 & 76.38 & 69.67 \\
        \midrule
        \rowcolor[HTML]{E3F2FD}
        \multirow{4}{*}{Trait Data} 
        & GenoAgent (Ours) & \textbf{46.81} & \textbf{63.71} & \textbf{32.28} \\
        & Round=1          & 45.04 & 59.25 & 30.74 \\
        & No Reviewer      & 24.02 & 32.58 & 6.45 \\
        & No Domain Expert & 25.14 & 23.48 & 4.68 \\
        \bottomrule
    \end{tabular}%
    \label{tab: preprocess}
\end{table}

\paragraph{Statistical analysis}
\label{sec:regression}
For the statistical analysis evaluation (Table~\ref{tab: regression}), we provided datasets preprocessed by human bioinformaticians and instructed various methods to perform statistical analysis following our standardized pipeline. Unlike data preprocessing, this task primarily involves applying standard Python libraries for statistical modeling, allowing several agent-based methods to achieve good performance. Fixed Python scripts implementing basic regression models like Lasso provided simple baselines but showed very low gene selection precision. Our comprehensive statistical analysis method—incorporating batch effect correction (BEC) and additional covariates—enabled more accurate identification of disease-related significant genes.

\subsection{Discussions}
While our results demonstrate the potential of LLM-based agentic methods for gene expression data analysis, they also highlight important limitations of current approaches.

\paragraph{Instability of the self-correction mechanism}
For complex tasks, the ideal scenario is for an agent to iteratively refine its code through feedback until reaching the correct solution. However, our results in Table \ref{tab: end2end_main} reveal that while a single round of self-correction significantly improves performance compared to no self-correction, additional rounds yield diminishing returns, leaving a substantial performance gap relative to human benchmarks. Our error analysis (Appendix \ref{sec: error_analysis}) demonstrates that the Code Reviewer agent often provides inconsistent feedback containing technical errors, which can mislead the programming agent rather than guide it toward improvement. We hypothesize this phenomenon stems from how LLMs are RLHF-tuned to prioritize following immediate instructions over maintaining consistent technical judgments across iterations. A promising research direction involves developing collaborative frameworks where agents can systematically discuss conflicting perspectives and resolve technical disagreements, potentially enhancing their collective understanding of complex tasks.

\paragraph{Limitations of the Domain Expert}
Clinical feature extraction emerged as the weakest link in our analysis pipeline. Despite the significant role of the domain expert agent, its performance still lagged behind human experts. Further investigation (Appendix \ref{sec: error_analysis}) showed that while the domain expert successfully isolated inference-heavy questions from broader task contexts, even the most advanced LLMs frequently made errors in specialized knowledge domains. The benefits of role-specific prompt design proved limited for these highly technical tasks. Future work could explore complementary techniques such as retrieval-augmented generation to address these knowledge limitations.

%% file: secs/con.tex
In this paper, we introduce GenoTEX, a benchmark for automated gene expression data analysis aimed at identifying disease-associated genes. The benchmark represents the standard pipeline in computational genomics, with problems and solutions curated by expert bioinformaticians. By structuring evaluation around three tasks—dataset selection, data preprocessing, and statistical analysis—GenoTEX provides a framework for assessing automated computational approaches to GTA analysis.

Our baseline method, GenoAgent, shows promising capabilities in gene expression analysis while revealing current limitations when handling biomedical data. The results highlight a performance gap between AI approaches and human experts, particularly in tasks requiring domain knowledge and clinical data interpretation.

GenoTEX provides researchers with a standardized resource for benchmarking and developing automated methods for genomic data analysis. Through its defined metrics and documentation, this benchmark aims to facilitate advances in AI-assisted genomics research, potentially leading to more efficient approaches to understanding genetic factors in human disease.

%% file: secs/ack.tex
This research was supported by the National AI Research Resource (NAIRR) under grant number 240283. We thank Yijiang Li for his suggestions to the baseline method, and Jinglin Jian, Jianrong Lu, Shuyi Guo, Taincheng Xing, Yuxuan Cheng, Jinglei Zhu, Mianchen Zhang, Miantong Zhang for their contribution to the benchmark curation.

%% file: secs/supplementary_materials.tex
\clearpage
\setcounter{section}{0} 
\onecolumn

\maketitlesupplementary

\renewcommand{\thesection}{\Alph{section}}

The supplementary material is organized as follows:

\begin{itemize}
\item Appendix \ref{sec: benchmark_detail} describes the dataset accessibility, documentation, and maintenance of our benchmark.
\item Appendix \ref{sec: guidelines} introduces the guidelines file used to standardize the manual curation of our benchmark.
\item Appendix \ref{sec: examples} provides examples of manual analysis on trait data extraction.
\item Appendix \ref{sec: criteria} outlines the criteria for forming trait-condition pairs for GTA analysis problems in our benchmark.
\item Appendix \ref{sec: data_acquisition} describes our data acquisition process.
\item Appendix \ref{sec: statistics} describes our statistical method for GTA analysis.
\item Appendix \ref{sec: attempts} presents our preliminary experiments highlighting the challenges faced by existing LLMs and agent-based methods on our benchmark.
\item Appendix \ref{sec: error_analysis} presents error analysis highlighting the limitations of GenoAgent on our benchmark and areas for future improvement.
\end{itemize}


\section{Dataset accessibility, documentation, and maintenance}
\label{sec: benchmark_detail}

\subsection{Documentation and intended uses}

GenoTEX is documented following the Datasheets for Datasets \cite{gebru2020datasheets} framework, providing a comprehensive description of the data collection process, preprocessing steps, and statistical analysis methods employed. The detailed datasheet is available \href{https://github.com/Liu-Hy/GenoTex/blob/main/datasheet.md}{here}. Each step of the pipeline aims to mirror the standards of computational genomics, ensuring that the dataset is both accurate and reliable. The intended uses of GenoTEX are broad, encompassing the evaluation and development of AI-driven methods for genomics data analysis. By providing a standardized benchmark, GenoTEX aims to facilitate the advancement of machine learning models capable of automating the complex task of gene expression analysis. Researchers in bioinformatics and related fields can leverage this dataset to benchmark their algorithms, fostering innovation and improving the scalability of gene-trait association (GTA) analysis processes.

\subsection{Open access and maintenance}

To ensure the accessibility and usability of GenoTEX, we have made the dataset publicly available \href{https://github.com/Liu-Hy/GenoTex}{here}. The dataset is hosted on GitHub, ensuring long-term availability and ease of access. The metadata associated with the dataset is documented \href{https://github.com/Liu-Hy/GenoTEX/blob/main/metadata.json}{here} using the Croissant Metadata Record \cite{Akhtar_2024}, providing a structured and detailed overview of the dataset's features and attributes. We have structured the metadata according to the JSON-LD standard \cite{json-ld-1.1} to enhance interoperability and organization. We will maintain the dataset with regular updates and ongoing support to address any issues or improvements that may arise. 

\subsection{Licensing and responsibility}

The GenoTEX dataset is released under a Creative Commons Attribution 4.0 International (CC BY 4.0) license \cite{cc-by-4.0}, which allows for broad usage while protecting the rights of the creators. The authors bear full responsibility for ensuring that the dataset adheres to this license and for any potential violations of rights. This responsibility includes ensuring that all data included in GenoTEX is ethically sourced and legally compliant. Throughout the curation process, we engaged in extensive discussions and consultations to address ethical considerations and legal requirements. Despite our best efforts, we remain aware that ethical landscapes can be complex and evolving, and we continually ask ourselves whether we are meeting the highest standards. This involved careful examination of each dataset to ensure the absence of personally identifiable information and compliance with all relevant standards. Our approach aims to ensure that GenoTEX meets the ethical and legal standards expected in the field of machine learning and computational genomics research.

\subsection{Data format and persistent identifiers}

GenoTEX is provided in open and widely used data formats, including CSV and JSON, ensuring compatibility with a wide range of analytical tools and platforms. Detailed instructions on how to read and use the dataset are included in the documentation, making it accessible to both novice and experienced researchers. To enhance the dataset's stability and ease of reference, we have created a DOI for GenoTEX: \url{https://doi.org/10.34740/kaggle/dsv/11309048}. This DOI will serve as a reliable means of access and citation for the dataset, promoting its use in academic and professional research.

\subsection{Reproducibility}

To support reproducibility, we have included all the necessary datasets, code, and evaluation procedures in our documentation. We have worked diligently to ensure that others can replicate the results of our analyses. By our commitment to transparency and reproducibility, we hope to facilitate the wider adoption and validation of AI-driven methods in genomics research.

\section{Guidelines for gene expression data analysis}
\label{sec: guidelines}
To tackle the complexities of gene expression data analysis, we have established a set of comprehensive guidelines shown below. These guidelines try to replicate the detailed processes of a skilled bioinformatician, covering dataset preprocessing, selection, and statistical analysis. By following these standardized procedures, we seek to improve consistency and reliability in our manual benchmark curation. 

\lstset{
  basicstyle=\ttfamily\small,
  frame=single,
  breaklines=true,
  breakindent=0pt,
  columns=fullflexible,
  keepspaces=true,
  showspaces=false,
  showstringspaces=false,
  showtabs=false,
  tabsize=2,
  captionpos=b,
  xleftmargin=0.02\textwidth, 
  xrightmargin=0.02\textwidth, 
  keywordstyle=\bfseries\color{keywords}, 
  commentstyle=\itshape\color{comments}, 
  keywords=[1]{Steps, Step, Notes}, 
}

\begin{lstlisting}[caption=Guidelines file for gene expression data analysis]
This document describes the standardized pipeline for analyzing gene expression data for identifying disease-associated genes, involving dataset preprocessing, selection, and statistical analysis. These steps follow the practices of computational genomics and ensure the reproducibility and reliability of the analysis. 

Data Sources and Organization:
- Gene expression data are sourced from two public databases, organized by trait in specific subdirectories:
    - Gene Expression Omnibus (GEO): Data are downloaded under certain criteria and saved under the path "{data_root}/GEO". Within this directory, datasets related to each trait are organized in subdirectories named after the trait.
    - The Cancer Genome Atlas (TCGA) data via the Xena platform: Data are saved under the path "{data_root}/TCGA". Similar to GEO, datasets related to each cancer type are organized in subdirectories named after the specific cancer trait.

Problem Setting Differentiation:
- If the problem is to identify significant genes predictive of a trait (optionally conditioning on age or gender, but not involving another trait), prepare the data related to this trait.
- If the problem is to identify significant genes predictive of a trait while conditioning on another trait, prepare data for both traits. These datasets will be integrated in a two-step regression process.

PART I. GEO Data Preprocessing

Step 1: Initial Data Loading
1. Identify the names of the SOFT file and Matrix file of the Series data.
2. Read the Matrix file to obtain background information and clinical trait data. This involves extracting the text data of series titles, summaries, and overall designs, as well as the tabular data of sample characteristics.
3. Get the unique values of all attributes in the sample characteristics table into a Python dictionary.
4. Print the background information and the sample characteristics dictionary for later observation.

Step 2: Dataset Analysis and Clinical Feature Extraction
1. Read the metadata to determine if the dataset is likely to contain gene expression data (which does not include miRNA data or methylation data).
2. Based on the metadata and the sample characteristics dictionary, for each of the variables of interest (e.g., a specific trait, age, gender):
    a. Assess the availability of data.
    b. If available, identify the key in the sample characteristics dictionary where unique values of this variable are recorded.
    c. Choose the appropriate data type (continuous or binary) and design conversion functions to encode the features into that type.
3. Conduct initial filtering. If either the gene data or trait data is not available, discard this dataset; otherwise, continue with the following steps.

Step 3: Gene Data Extraction
1. Read the Matrix file to extract the tabular gene expression data into a dataframe.
2. Print the first few row identifiers in the dataframe for later observation.
3. Determine if the row identifiers are human gene symbols or other types that require mapping.

Step 4: Gene Annotation (Conditional)
1. If gene mapping is required, extract the gene annotation table from the SOFT file.
2. Preview the gene annotation table for later observation.

Step 5: Gene Identifier Mapping
1. If gene mapping is required, identify the columns for the identifiers and gene symbols from the gene annotation table.
2. Create a mapping dataframe and apply it to the gene expression data. Handle many-to-many relationships between probe IDs and gene symbols by splitting concatenated strings of symbols using separators such as semicolons (;), vertical bars (|), double slashes (//), and commas (,). Assign the corresponding expression values to each gene symbol linked to an identifier. Finally, aggregate the expression values for each gene symbol by averaging the values from multiple probes, with the aim of accurately representing the expression level of each gene symbol.

Step 6: Data Normalization and Merging
1. Normalize the gene symbols in the gene data by querying databases with the Python MyGene library, setting the `scopes' parameter properly. Remove data corresponding to genes that cannot be normalized. For genes that normalize to the same symbol, deduplicate by averaging their expression values.
2. Merge the clinical data with the normalized gene data on sample IDs.
3. Handle missing values. Drop records with the clinical trait missing or with more than 20% of the gene features missing. Use mean imputation for other missing values in the gene expression data.
4. Observe the resulting dataset for quality verification. If the dataset is successfully preprocessed, save the merged data to a CSV file.

PART II. TCGA-Xena Data Preprocessing

Step 1: Initial Data Loading
1. Identify the names of the clinical data file and the genetic data file, and load them into two separate dataframes. For gene expression, we choose the `gene expression RNAseq' dataset instead of its PANCAN normalized or percentile versions.

Step 2: Clinical Attribute Selection
1. Print and observe the column names of the clinical data file. Identify all columns that might hold relevant data for age and gender from the list of column names.
2. Inspect the first few values of all candidate columns. Select a single column from the candidate columns that accurately records age and gender information, respectively, considering meaningful values and minimal missing data.
3. Based on metadata of the TCGA database, use a simple rule to convert the trait (whether the sample has the particular type of cancer) to binary values.
4. Conduct initial filtering. If all samples have the same target values, or if the clinical dataset shows other abnormalities, discard the dataset. Otherwise, continue with the next step.

Step 3: Data Processing and Merging
1. Normalize the gene symbols in the gene data by querying databases with the Python MyGene library, setting the `scopes' parameter properly. Remove data corresponding to genes that cannot be normalized. For genes that normalize to the same symbol, deduplicate by averaging their expression values.
2. Merge the clinical and genetic datasets on sample IDs.
3. Handle missing values. Drop records with the clinical trait missing or with more than 20% of the gene features missing. Use mean imputation for other missing values in the gene expression data.
4. Observe the resulting dataset for quality verification. If the dataset is successfully preprocessed, save the merged data to a CSV file.

PART III. Statistical Analysis

Step 1: Data Selection and Loading
1. Select the best input data relevant to the GTA analysis problem, and load the data into a dataframe. If multiple preprocessed datasets are available for statistical analysis about a trait, we select the one with the largest sample size.
2. If the analysis requires integrating datasets about two traits, we sort the possible pairs of datasets for both traits by the product of their sample sizes, and select the pair with the largest product. Load data for the trait and condition into separate dataframes and select common gene regressors.

Step 2: Data Wrangling
1. Extract the relevant data columns and convert into numpy arrays for analysis. Get the data matrices of features, the target variable, and also the condition when applicable.
2. For two-step regression, this needs to be done twice. In the first step, the features are the common gene regressors, and the target is the condition, and we need to extract these matrices from the condition dataset. The second step follows other cases for extracting relevant data.

Step 3: Condition Prediction (Only for Two-Step Regression)
1. Determine the variable type (continuous or binary) of the condition.
2. Select a simple regression model based on the type of the target variable, and train it to regress the condition on the common gene regressors in the condition dataset.
3. Use the trained model to predict the condition values in the trait dataset using the common gene regressors. Remove the columns in the trait dataset corresponding to the common regressors, and add the predicted condition values to it as a new column.

Step 4: Model Selection Based on Batch Effect
1. Assess whether the dataset shows batch effects by observing gaps in eigenvalues. Choose the appropriate model based on the presence of batch effects. Use a Linear Mixed Model (LMM) if batch effects are detected. Otherwise, use a Lasso model.

Step 5: Data Normalization
1. For the feature matrix, and the condition matrix (if applicable), apply Z-score normalization so that each feature has a mean of 0 and standard deviation of 1. Make sure this is done every time before training the model.

Step 6: Hyperparameter Tuning
1. Do 5-fold cross-validation, and perform hyperparameter search on the logarithm scale with base of 10. Record the best hyperparameter settings.

Step 7: Model Training
1. Train the model on the entire dataset, with the best hyperparameters found during cross-validation. For conditional analyses, incorporate the condition matrix into the model.

Step 8: Model Interpretation
1. Interpret the trained model to identify significant factors and effects. For Lasso, choose gene variables with non-zero coefficients. For LMM, apply the Benjamini-Hochberg correction for multiple hypothesis testing, and select variables whose corrected p-value is less than 0.05.
2. Save the regression output to a JSON file, with the identified genes and the corresponding coefficient or p-values.
\end{lstlisting}

\lstset{
  basicstyle=\ttfamily\small,
  frame=single,
  breaklines=true,
  breakindent=0pt,
  columns=fullflexible,
  keepspaces=true,
  showspaces=false,
  showstringspaces=false,
  showtabs=false,
  tabsize=2,
  captionpos=b,
  xleftmargin=0.02\textwidth, 
  xrightmargin=0.02\textwidth, 
  keywordstyle=\bfseries\color{blue}, 
  commentstyle=\itshape\color{gray}, 
  stringstyle=\color{green}, 
  identifierstyle=\color{black}, 
  keywordstyle=[1]\color{blue}, 
  keywordstyle=[2]\color{purple}, 
}

\section{Examples of manual analysis}
\label{sec: examples}
In addition to the guidelines file, we provide example files to the participants of our benchmark curation. These examples include code and results for analyzing GTA analysis problems related to traits such as \textit{Breast Cancer} and \textit{Epilepsy}. These illustrations have proven helpful in familiarizing participants with these tasks quickly. Among the many steps in the analysis pipeline, a key step is the trait data extraction during the preprocessing of GEO data. This step requires biomedical knowledge and an understanding of the dataset collection process described in the metadata. In this section, we will introduce the part of the manual analysis examples related to this crucial step.

\subsection{Problem statement}
Our goal was to extract clinical traits from GEO datasets. For each trait of interest, we aimed to determine its availability and develop encoding rules to automate the extraction process. Below are two examples focusing on \textit{Breast Cancer} and \textit{Epilepsy}, respectively.

\subsection{Breast Cancer example}
\subsubsection{Input data}
\begin{lstlisting}[caption=Background information for breast cancer]
!Series_title    "Unlocking Molecular mechanisms and identifying druggable targets in matched-paired brain metastasis of Breast and Lung cancers"
!Series_summary  "Introduction: The incidence of brain metastases in cancer patients is increasing, with lung and breast cancer being the most common sources. Despite advancements in targeted therapies, the prognosis remains poor, highlighting the importance to investigate the underlying mechanisms in brain metastases. The aim of this study was to investigate the differences in the molecular mechanisms involved in brain metastasis of breast and lung cancers. In addition, we aimed to identify cancer lineage-specific druggable targets in the brain metastasis. Methods: To that aim, a cohort of 44 FFPE tissue samples, including 22 breast cancer and 22 lung adenocarcinoma (LUAD) and their matched-paired brain metastases were collected. Targeted gene expression profiles of primary tumors were compared to their matched-paired brain metastases samples using nCounter PanCancer IO 360 Panel of NanoString technologies. Pathway analysis was performed using gene set analysis (GSA) and gene set enrichment analysis (GSEA). The validation was performed by using Immunohistochemistry (IHC) to confirm the expression of immune checkpoint inhibitors. Results: Our results revealed the significant upregulation of cancer-related genes in primary tumors compared to their matched-paired brain metastases (adj. p<=0.05). We found that upregulated differentially expressed genes in breast cancer brain metastasis (BM-BC) and brain metastasis from lung adenocarcinoma (BM-LUAD) were associated with the metabolic stress pathway, particularly related to the glycolysis. Additionally, we found that the upregulated genes in BM-BC and BM-LUAD played roles in immune response regulation, tumor growth, and proliferation. Importantly, we identified high expression of the immune checkpoint VTCN1 in BM-BC, and VISTA, IDO1, NT5E, and HDAC3 in BM-LUAD. Validation using immunohistochemistry further supported these findings. Conclusion: In conclusion, the findings highlight the significance of using matched-paired samples to identify cancer lineage-specific therapies that may improve brain metastasis patients outcomes."
!Series_overall_design    "RNA was extracted from FFPE samples of (primary LUAD and their matched paired brain metastasis n=22, primary BC and their matched paired brain metastasis n=22)"
\end{lstlisting}

\begin{lstlisting}[caption=Sample characteristics for breast cancer. Some long lists are truncated for brevity.]
{
  0: ['age at diagnosis: 49', 'age at diagnosis: 44', 'age at diagnosis: 41', 'age at diagnosis: 40', ...],
  1: ['Sex: female', 'Sex: male'],
  2: ['histology: TNBC', 'histology: ER+ PR+ HER2-', 'histology: Unknown', 'histology: ER- PR- HER2+', 'histology: ER+ PR-HER2+', 'histology: ER+ PR- HER2-', 'histology: ER- PR+ HER2-', 'histology: adenocarcinoma'],
  3: ['smoking status: n.a', 'smoking status: former-smoker', 'smoking status: smoker', 'smoking status: Never smoking', 'smoking status: unknown', 'smoking status: former-roker'],
  4: ['treatment after surgery of bm: surgery + chemotherpy', 'treatment after surgery of bm: surgery + chemotherpy + Radiotherapy', 'treatment after surgery of bm: surgery + chemotherapy + Radiotherapy', 'treatment after surgery of bm: surgery', 'treatment after surgery of bm: surgery + chemotherapy + Radiotherapy', ...]
}
\end{lstlisting}

\subsubsection{Inference process}

The dataset summary indicated that tissue samples from primary breast cancer (BC) and lung adenocarcinoma (LUAD), along with their matched-paired brain metastases, were included. By examining the sample characteristics dictionary, combined with domain knowledge, we identified subtypes such as 'TNBC', 'ER+', 'PR+', and 'HER2+' associated with breast cancer, and 'adenocarcinoma' associated with lung cancer. Based on this, we developed a rule: tissues labeled with 'TNBC', 'ER+', 'PR+', or 'HER2+' are coded as having breast cancer (1), while 'adenocarcinoma' is coded as not having breast cancer (0).

\begin{lstlisting}[language=Python, caption=Python function to encode Breast Cancer trait]
def convert_trait(value):
    if 'TNBC' in value or 'ER+' in value or 'PR+' in value or 'HER2+' in value:
        return 1  # Breast Cancer
    elif 'adenocarcinoma' in value:
        return 0  # Not Breast Cancer (LUAD)
    else:
        return None  # Unknown
\end{lstlisting}

\subsection{Epilepsy example}

\subsubsection{Input data}

\begin{lstlisting}[caption={Background information for Epilepsy}]
!Series_title    "Integrated analysis of expression profile and potential pathogenic mechanism of temporal lobe epilepsy with hippocampal sclerosis"
!Series_summary  "To investigate the potential pathogenic mechanism of temporal lobe epilepsy with hippocampal sclerosis (TLE+HS), we have employed analyzing of the expression profiles of microRNA/ mRNA/ lncRNA/ DNA methylation in brain tissues of hippocampal sclerosis (TLE+HS) patients. Brain tissues of six patients with TLE+HS and nine of normal temporal or parietal cortices (NTP) of patients undergoing internal decompression for traumatic brain injury (TBI) were collected. The total RNA was dephosphorylated, labeled, and hybridized to the Agilent Human miRNA Microarray, Release 19.0, 8x60K. The cDNA was labeled and hybridized to the Agilent LncRNA+mRNA Human Gene Expression Microarray V3.0, 4x180K. For methylation detection, the DNA was labeled and hybridized to the Illumina 450K Infinium Methylation BeadChip. The raw data was extracted from hybridized images using Agilent Feature Extraction, and quantile normalization was performed using the Agilent GeneSpring. We found that the disorder of FGFR3, hsa-miR-486-5p, and lnc-KCNH5-1 plays a key vital role in developing TLE+HS."
!Series_overall_design    "Brain tissues of six patients with TLE+HS and nine of normal temporal or parietal cortices (NTP) of patients undergoing internal decompression for traumatic brain injury (TBI) were collected."
\end{lstlisting}

\begin{lstlisting}[caption=Sample characteristics for Epilepsy]
{
  0: ['tissue: Hippocampus', 'tissue: Temporal lobe', 'tissue: Parietal lobe'],
  1: ['gender: Female', 'gender: Male'],
  2: ['age: 23y', 'age: 29y', 'age: 37y', 'age: 26y', 'age: 16y', 'age: 13y', 'age: 62y', 'age: 58y', 'age: 63y', 'age: 68y', 'age: 77y', 'age: 59y', 'age: 50y', 'age: 39y']
}
\end{lstlisting}

\subsubsection{Inference process}

The dataset summary indicated that brain tissues from patients with temporal lobe epilepsy with hippocampal sclerosis (TLE+HS) and control samples were included. By examining the sample characteristics dictionary, we identified tissue types such as 'Hippocampus', 'Temporal lobe', and 'Parietal lobe'. We inferred that 'Hippocampus' and 'Temporal lobe' tissues are associated with TLE+HS (epilepsy), while 'Parietal lobe' tissues are from control samples. Based on this, we developed a rule: tissues labeled with 'Hippocampus' or 'Temporal lobe' are coded as having epilepsy (1), while 'Parietal lobe' is coded as control (0).

\begin{lstlisting}[language=Python, caption=Python function to encode Epilepsy trait]
def convert_trait(value):
    if 'Hippocampus' in value or 'Temporal lobe' in value:
        return 1  # Epilepsy (TLE+HS)
    elif 'Parietal lobe' in value:
        return 0  # Control (NTP)
    else:
        return None  # Unknown
\end{lstlisting}

\subsection{Validation and conclusion}

By executing the provided Python functions, we confirmed the accuracy of our trait extraction process. For instance, applying the \texttt{convert\_trait} function for the epilepsy dataset, we verified the presence of exactly six samples with the positive \textit{Epilepsy} trait, consistent with the metadata description. Similarly, for the breast cancer dataset, the function accurately identified 22 samples with the \textit{Breast Cancer} trait. These examples highlight the dataset context understanding and domain knowledge inference required for the accurate preprocessing of gene expression data. 

\section{Criteria for manual correction of trait-condition pairs}
\label{sec: criteria}

To ensure the scientific validity of our benchmark problems, we apply specific rules for including and excluding certain trait-condition pairs. Each biomedical entity in our list can be considered a trait and paired with a condition, where the condition is either another entity from the list or a demographic attribute like ``age'' or ``gender''. The following criteria are designed to maintain scientific relevance and robustness:

\begin{itemize}
    \item \textbf{Trait-Condition Role Assignment}: Entities such as Vitamin D levels and LDL cholesterol levels are included only as conditions and not as traits. This distinction ensures that the primary focus remains on traits with more direct clinical implications, while these entities serve as influential factors that could affect those traits.

    \item \textbf{Universal Conditions}: Entities such as obesity and hypertension are designated as conditions to be paired with all other traits. This is because these conditions are widespread and significantly impact various health outcomes, making them critical factors to consider in any genetic analysis.

    \item \textbf{Gender-Specific Considerations}: Gender-specific entities such as endometriosis and prostate cancer are not conditioned on gender. Furthermore, entities from different genders are not paired. This approach respects the biological distinctions between genders and ensures that the resulting problems remain relevant and meaningful.

    \item \textbf{Cancer Category Exclusion}: Pairs where both the trait and the condition belong to the cancer category are excluded. This is because investigating genetic factors behind one type of cancer conditioned on another type of cancer is often less scientifically important. The focus is placed on broader, more impactful genetic relationships that offer greater insight into cancer biology.
\end{itemize}

These criteria are used in combination with the Jaccard similarity of associated genes (Section \ref{sec: benchmark_creation}), to ensure the biological significance of the benchmark problems.

\section{Details about data acquisition}
\label{sec: data_acquisition}

Our benchmark required two categories of data sources: (1) gene expression and clinical datasets as primary input data, and (2) domain knowledge repositories to support analysis and evaluation. Below, we detail our systematic approach to acquiring and processing data from each source.

\subsection{Input Gene Expression Data Sources}

\paragraph{Gene Expression Omnibus (GEO)} GEO \citep{clough2016gene} is a comprehensive public repository for high-throughput genomic data. We implemented a systematic acquisition protocol using the Entrez Programming Utilities to search GEO DataSets (a manually curated subset of GEO) for human gene expression series relevant to each trait in our study. Our selection process followed these criteria:

\begin{enumerate}
  \item \textbf{Search parameters:}
  \begin{itemize}
    \item Sample size: Between 30 and 10,000 subjects
    \item Organism: Homo sapiens (human)
    \item Publication period: 2010-2025
    \item Data types: Expression profiling (array and high-throughput sequencing), genome variation profiling (high-throughput sequencing and SNP array), SNP genotyping, and third-party reanalysis
  \end{itemize}
  
  \item \textbf{File requirements:}
  \begin{itemize}
    \item Presence of both SOFT (Series and Platform family) files and matrix files
    \item Matrix file size: 0.1MB-100MB (ensuring sufficient gene expression data while maintaining computational tractability)
    \item Family file size: Less than 100MB
  \end{itemize}
  
  \item \textbf{Cohort selection:}
  \begin{itemize}
    \item For each trait, we selected up to 10 cohort datasets that satisfied all criteria
    \item When more than 10 datasets were available, we used GEO's default ranking system (which prioritizes recency and relevance)
  \end{itemize}
\end{enumerate}

This approach ensured a balanced representation of diverse, high-quality datasets of manageable size for each trait.

\paragraph{The Cancer Genome Atlas (TCGA)} The TCGA \citep{tomczak2015cancer} provides standardized gene expression and clinical data for multiple cancer types. We accessed these datasets through the UCSC Xena platform \citep{ucsc_xena}, which offers harmonized data in a consistent format. Unlike our selective approach with GEO, we comprehensively incorporated all available TCGA datasets related to 36 cancer types for several reasons:

\begin{enumerate}
  \item \textbf{Standardized format:} TCGA data through Xena has consistent processing, with uniformly formatted gene expression matrices and clinical information
  \item \textbf{High quality:} The datasets undergo rigorous quality control through both TCGA and the Xena platform
  \item \textbf{Integrated patient IDs:} Expression and clinical data are consistently linked through standardized patient identifiers
  \item \textbf{Appropriate size:} Dataset sizes are consistently suitable for computational analysis without additional filtering
\end{enumerate}

This comprehensive inclusion of TCGA data complemented our more selective approach to GEO datasets, providing our benchmark with standardized cancer genomics data.

\subsection{Domain Knowledge Sources}

\paragraph{NCBI Gene Database} To address nomenclature inconsistencies across expression datasets, we utilized the NCBI Gene database \citep{NCBIGene} to standardize gene symbols. Our process involved:

\begin{enumerate}
  \item Extracting official human gene symbols and their synonyms from the NCBI Gene FTP repository
  \item Creating a gene synonym dictionary mapping all gene aliases (converted to uppercase) to their respective official symbols
  \item Implementing precedence rules to resolve ambiguities where multiple official symbols share the same synonym
\end{enumerate}

This standardization facilitated the integration of expression data from heterogeneous sources and ensured consistent gene identification across all datasets in our benchmark.

\paragraph{Open Targets Platform} The Open Targets Platform \citep{opentargets} integrates multi-omics data to identify and prioritize disease-associated targets. We specifically utilized the ``associationByOverallDirect'' dataset, which contains direct target-disease associations with overall scores aggregated across all evidence types. Our extraction process included:

\begin{enumerate}
  \item Downloading the association dataset from Open Targets using their rsync service
  \item Mapping each trait name to a disease ID using the Open Targets API
  \item Filtering associations to include only those with confidence scores exceeding 0.2 to ensure biological relevance
  \item Converting target Ensembl IDs to gene symbols using mygene.info \cite{wu2013biogps} and normalizing them using our gene synonym dictionary
\end{enumerate}

These curated gene-disease associations served two crucial functions in our benchmark construction:

\begin{enumerate}
  \item \textbf{Selection of meaningful trait-condition pairs:} We calculated Jaccard similarity between trait-related and condition-related gene sets, selecting pairs with similarity exceeding 0.1 for their shared biological mechanisms
  \item \textbf{Identification of biologically-relevant regressors:} We identified genes common between trait and condition datasets that also had known associations with the condition according to Open Targets. This approach enabled us to train regression models on condition datasets and apply them to predict condition status in trait datasets, facilitating systematic study of how specific conditions influence gene-trait relationships. Section \ref{sec: statistics} provides more details on this ``two-step regression'' approach.
\end{enumerate}

\section{Details of statistical method for GTA analysis}
\label{sec: statistics}
\textbf{Notations}
We consider $\X \in \mathbb{R}^{n \times p}$ as the data matrix,
where $n$ represents the number of samples, 
and $p$ the number of gene expressions. 
We will use $\y \in \mathbb{R}^n$ to denote the disease status.
The basic data generation process is assumed to be 
\begin{align}
    \y = \X\beta + \epsilon,
    \label{eq:datagen}
\end{align}
where we use $\beta$ to denote the coefficients to estimate and $\epsilon \sim N(\mathbf{0}, \mathbf{I}\sigma_\epsilon^2)$. 

\textbf{Vanilla Case}
In the most vanilla case, Lasso is chosen for its effectiveness in variable selection within high-dimensional datasets, isolating the most informative genes while controlling for overfitting. 
Lasso \cite{tibshirani1996regression} addresses this task by solving:
\begin{align}
\min_{\beta} \frac{1}{2}\Vert \y - \X\beta\Vert_2^2 +  \lambda \Vert\beta\Vert,
\label{eq:lasso}
\end{align}
where $\beta$ is the coefficients to estimate (non-zero coefficients mean that Lasso selects these genes), 
and $\lambda$ is a hyperparameter. 

\textbf{Confounding Factor Detection}
The above Vanilla case is applied when there are no confounding factors in the data. 
However, 
before that, we need first check the data to inspect 
whether there are confounding factors. 
Inspired by the Marchenko-Pastur Law \cite{johnstone2001distribution}
and 
following a previous heuristic \cite{wang2017variable}, 
we inspect the ``gap'' of eigenvalues of the covariance matrix 
$\X\X^T$. 
If there is a clear gap in the eigenvalues, 
we consider the data are from multiple distributions, 
and thus, potentially with confounding factors. 
More formally, 
the confounding factor detection will return true 
if $e_i - e_{i+1} > 1/n$ holds for any $i=0, 1, \dots, t$, where $e_i$ stands for the $i$\textsuperscript{th} eigenvalues if eigenvalues have been sorted descendingly, and $t$ stands for the maximum number of eigenvalues we inspect. 

After the confounding factor detection, 
if we observe no confounding factors in the data, 
we directly follow the vanilla case above. 
However, if there are confounding factors, 
we adopt one of the following two approaches. 
\begin{itemize}
    \item Regression with Confounding Factor Correction
    \item Regression after Confounding Factor Correction
\end{itemize}

\textbf{Regression with Confounding Factor Correction}
The first choice is to take advantage of the impressive empirical power 
of linear mixed model in correct confounding factors for genetic association studies \cite{yu2006unified,lippert2011fast,wang2022trade}.
With the setup of linear mixed model, we can formulate the problem with
\begin{align}
    \y = \X\beta + \Z\bu + \epsilon, 
    \label{eq:lmm}
\end{align}
where $\Z$ is an indication of the underlying population, and $\bu \sim N(\mathbf{0}, \mathbf{I}\sigma_u^2)$ and $\epsilon \sim N(\mathbf{0}, \mathbf{I}\sigma_\epsilon^2)$, the dimension of $\mathbf{0}$ and $\mathbf{I}$ can be inferred from the context, and $\sigma$ denotes the standard deviation of the Gaussian distribution. 

We use several tricks to estimate $\beta$ and the solution for a single column of $\X$ (denoted by $\x$) is 
\begin{align*}
    \beta_\x = \dfrac{\x^T(\widehat{\delta}\mathbf{I} + \X\X^T)^{-1}\y}{\x^T(\widehat{\delta}\mathbf{I} + \X\X^T)^{-1}\x},
\end{align*}
because the trick will use $\X$ to replace the unknown $\Z$ \cite{lippert2011fast,wang2022trade}. Also, we use $\widehat{\delta}$ to denote the estimated ratio of the two $\sigma$. 

Note that linear mixed model will estimate the coefficients individually for each gene, which is not necessarily ideal for our analysis. As an alternative choice, we can also use the following. 

\textbf{Regression after Confounding Factor Correction}
In order to have a multivariate method, where the coefficients of genes can be considered together, 
we can also choose to correct the confounding factors first. 
The correction is mostly built upon the theory of how linear mixed model
helps to correct confounding factors \cite{wang2022trade}, 
and achieved by the rotation of $\X$ as following:
\begin{align*}
    \Tilde{\X} = (\widehat{\delta}\mathbf{I} + \X\X^T)^{-\frac{1}{2}}\X
\end{align*}
where $\widehat{\delta}$ is again the estimation ratio of the two sigmas.

With this rotation, one can perform Lasso in the vanilla case again 
with $\Tilde{\X}$ and $\y$ for estimation of coefficients. 

\textbf{Incorporation of Conditions}
To make sure the regression analysis will incorporate different conditions such as age and gender, 
we will update our basic assumption of the data generation of Eq.~\ref{eq:datagen} into 
\begin{align}
    \y = \X\beta + \C\alpha + \epsilon 
    \label{eq:datagen2}
\end{align}
where we use $\C$ to denote additional conditions such as age and gender, 
and use $\alpha$ to denote the corresponding coefficient effect sizes. 

Note that Eq.~\ref{eq:datagen2} is to replace Eq.~\ref{eq:datagen} as an underlying data generation assumption. 
All the above solutions to estimate $\beta$ for Eq.~\ref{eq:datagen} still apply to Eq.~\ref{eq:datagen2} by incorporating $\C$ into Eq.~\ref{eq:lasso} or Eq.~\ref{eq:lmm}.  

\textbf{Two-step Regression}
While the above process offers a solution to incorporate 
the conditions when they are known. 
There are many cases that we do not necessarily have the conditions in the data. 
In this case, we will follow the following procedure, 
which we name two-step regression. 
It uses common genes between datasets that are known for their association with the condition, 
to estimate the conditions that our analyses need but do not know.

For example, we are interested in studying the association between disease $d$ and gene expressions given conditions $c$. However, there are no gene expression samples collected with both $d$ and $c$ appear in the same dataset. 
In this case, we will identify two datasets: Dataset 1 has $\y^{(1)}$ of disease $d$ and the corresponding gene expression $\X^{(1)}$, 
and Dataset 2 has $\y^{(2)}$ of the condition $c$ and the corresponding gene expression $\X^{(2)}$. 
We also need to identify a common set of genes $\X_{c}^*$ that appear in both datasets. 
In addition, 
the selection criteria for $\X_{c}^*$ also include their established biomedical relevance to the condition.

In our two-step regression, 
the first step is to construct the following regression model
\begin{equation}
    \y^{(2)} = \X^{(2)}_{c}\beta_{c} + \epsilon_c
\end{equation} 
$\beta_{c}$ are the coefficients indicating the effects of these genes on the condition, and $\epsilon_c$ is the error term. 

In the second step, 
we first predict the missing conditions for the first dataset, with
\begin{align*}
    \C = \X^{(2)}_{c}\widehat{\beta}_{c}, 
\end{align*}
where $\widehat{\beta}_{c}$ is the estimated $\beta_{c}$ from the first step. 

Then, we create a $\X^{(2)}_t$ by excluding the column of $\X^{(2)}_{c}$ out of $\X^{(2)}$. 
Then the remaining follows by performing the analysis with $\y$, 
$\X^{(2)}_t$, and $\C$.

\section{Challenges faced by existing methods on our benchmark}
\label{sec: attempts}
Gene expression data analysis is a complex and specialized task. Despite their problem-solving abilities, state-of-the-art LLMs and agent-based methods struggle with gene expression data. Our evaluations of methods such as GPT-4o \cite{openai2024gpt4o}, MetaGPT \cite{hong2023metagpt}, and CodeAct \cite{wang2024executable} revealed consistent failures across various settings.

We tested these methods under three different settings: (i) providing general task instructions, (ii) providing detailed task instructions used by GenoAgent, and (iii) providing detailed task instructions and all necessary library functions as in GenoAgent. Each setting was tested on a subset of 50 GTA analysis problems. Our results show that none of the methods generated runnable code for preprocessing datasets downloaded from GEO. Persistent errors in the generated code prevented testable outputs, regardless of the level of detail provided.

First, we find that when preprocessing GEO data, these methods often fail at dataset loading in the initial steps. The gene expression data files follow special formats. The agent struggles to extract tabular data embedded in the text file by identifying special markers, skipping metadata rows, and setting other parameters correctly, resulting in data reading failures.

\begin{lstlisting}[language=Python, caption=Failure example of MetaGPT in reading datasets]
import pandas as pd
from typing import Tuple
from utils import Utils

class DataLoader:
    """
    DataLoader class is responsible for loading clinical and genetic data from given file paths.
    """

    def load_clinical_data(self, filepath: str) -> pd.DataFrame:
        """
        Loads clinical data from a specified file path.

        :param filepath: The path to the clinical data file.
        :return: A pandas DataFrame containing the clinical data.
        """
        try:
            clinical_data = pd.read_csv(filepath)
            Utils.log(f"Clinical data loaded successfully from {filepath}")
            return clinical_data
        except FileNotFoundError:
            Utils.log(f"File not found: {filepath}")
            raise
        except pd.errors.EmptyDataError:
            Utils.log(f"No data: {filepath} is empty")
            raise
        except Exception as e:
            Utils.log(f"An error occurred while loading clinical data: {e}")
            raise
\end{lstlisting}

We manually corrected the data loading code for the baseline methods and continued with the tasks. However, they were still unable to conduct the inference required to extract clinical features. This step is inherently difficult and often requires at least one round of debugging by the Domain Expert agent in our GenoAgent method to achieve a higher success rate.
\begin{lstlisting}[language=Python, caption=Failure example of MetaGPT in encoding Breast Cancer trait]
def convert_trait(self, value: str) -> str:
    """
    Converts a trait value to a standardized string format.

    :param value: The trait value to convert.
    :return: A standardized string representation of the trait.
    """
    # This is a placeholder for the actual conversion logic, which would
    # depend on the specific requirements for trait conversion.
    # For example, it could map various synonyms to a canonical form.
    standardized_value = value.strip().lower()
    return standardized_value
\end{lstlisting}

\begin{lstlisting}[language=Python, caption=Failure example of CodeAct in encoding Breast Cancer trait. 'TLE+HS' is indeed related to epilepsy according to the metadata\texttt{,} but this is not the way the trait information is recorded for each sample. Moreover\texttt{,} these functions didn't strip the content before the colon. As a result\texttt{,} the code will convert all trait values to None.]
def convert_trait(value):
    if value in ['TLE+HS', 'control']:
        return 1 if value == 'TLE+HS' else 0
    return None
\end{lstlisting}

The challenges faced by methods like MetaGPT and CodeAct in processing gene expression data primarily stem from their difficulty in handling specialized data formats and the absence of flexible feedback mechanisms. MetaGPT, primarily designed for software engineering tasks, operates with an independent execution model and limited context-awareness, which can impede dynamic adaptation during task execution and lead to errors when dealing with the nuanced formats of gene expression datasets. CodeAct, while effective at generating executable code through structured prompts, lacks the context-aware planning and iterative refinement necessary for the intricate steps involved in gene expression data preprocessing. Its static approach does not easily accommodate the dynamic adjustments required for diverse and complex gene expression data, leading to errors during initial data loading and clinical feature extraction.

In contrast, GenoAgent employs a team of specialized agents that maintain a comprehensive task context and leverage expert consultation, allowing for context-aware planning and iterative correction. This enables GenoAgent to handle the complexities of genomics data analysis more effectively, improving its reliability in data preprocessing.

\section{Discussion on the limitations of GenoAgent}
\label{sec: error_analysis}

\begin{figure}[htbp]
\centering
\begin{minipage}[t]{0.48\textwidth}
\centering
\includegraphics[width=1.0\textwidth]{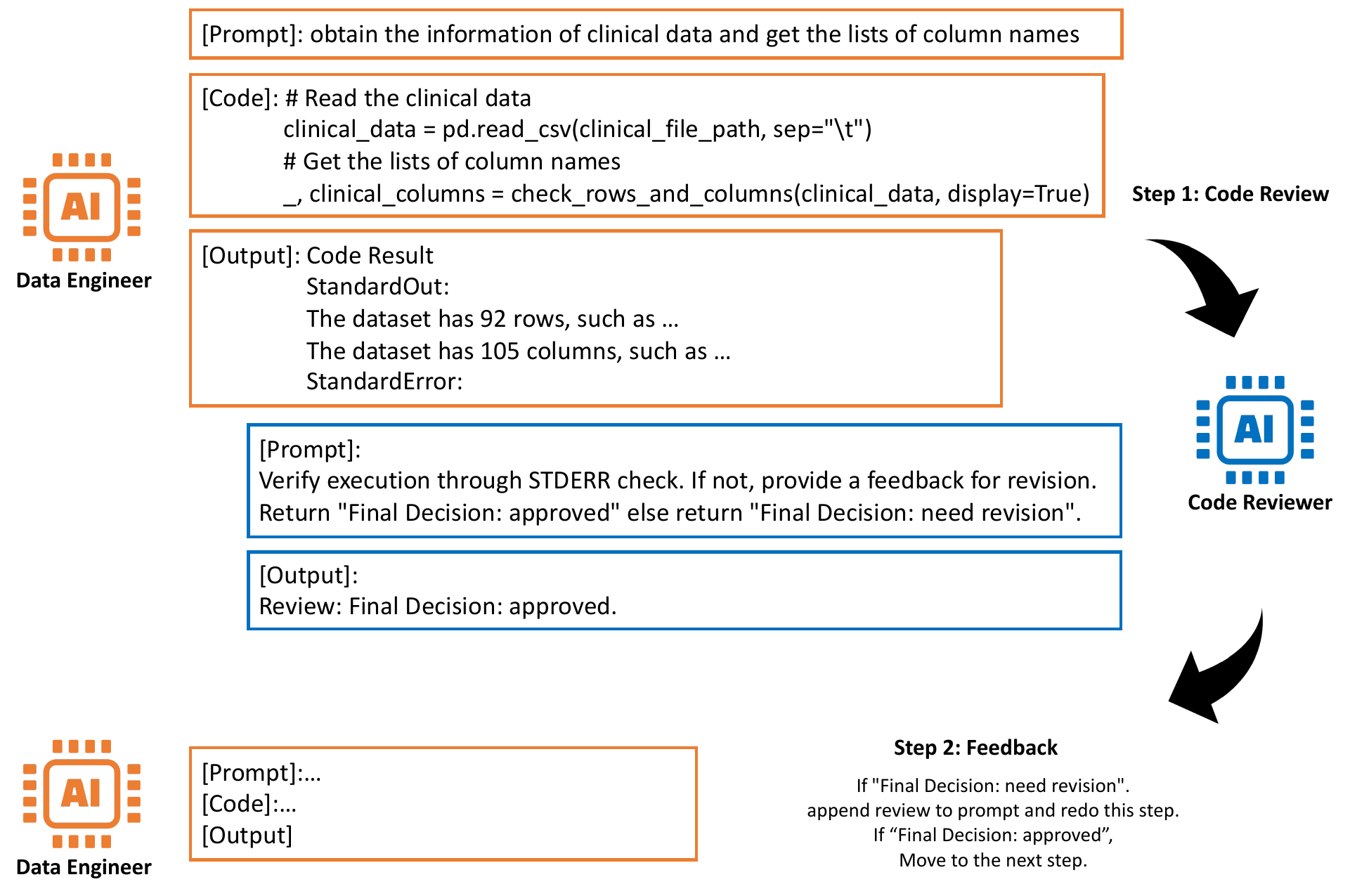}
    \caption{The collaboration between Data Engineer and Code Reviewer. 
    }
    \label{fig:reviewer}
\end{minipage}
\begin{minipage}[t]{0.48\textwidth}
\centering
\includegraphics[width=1.0\textwidth]{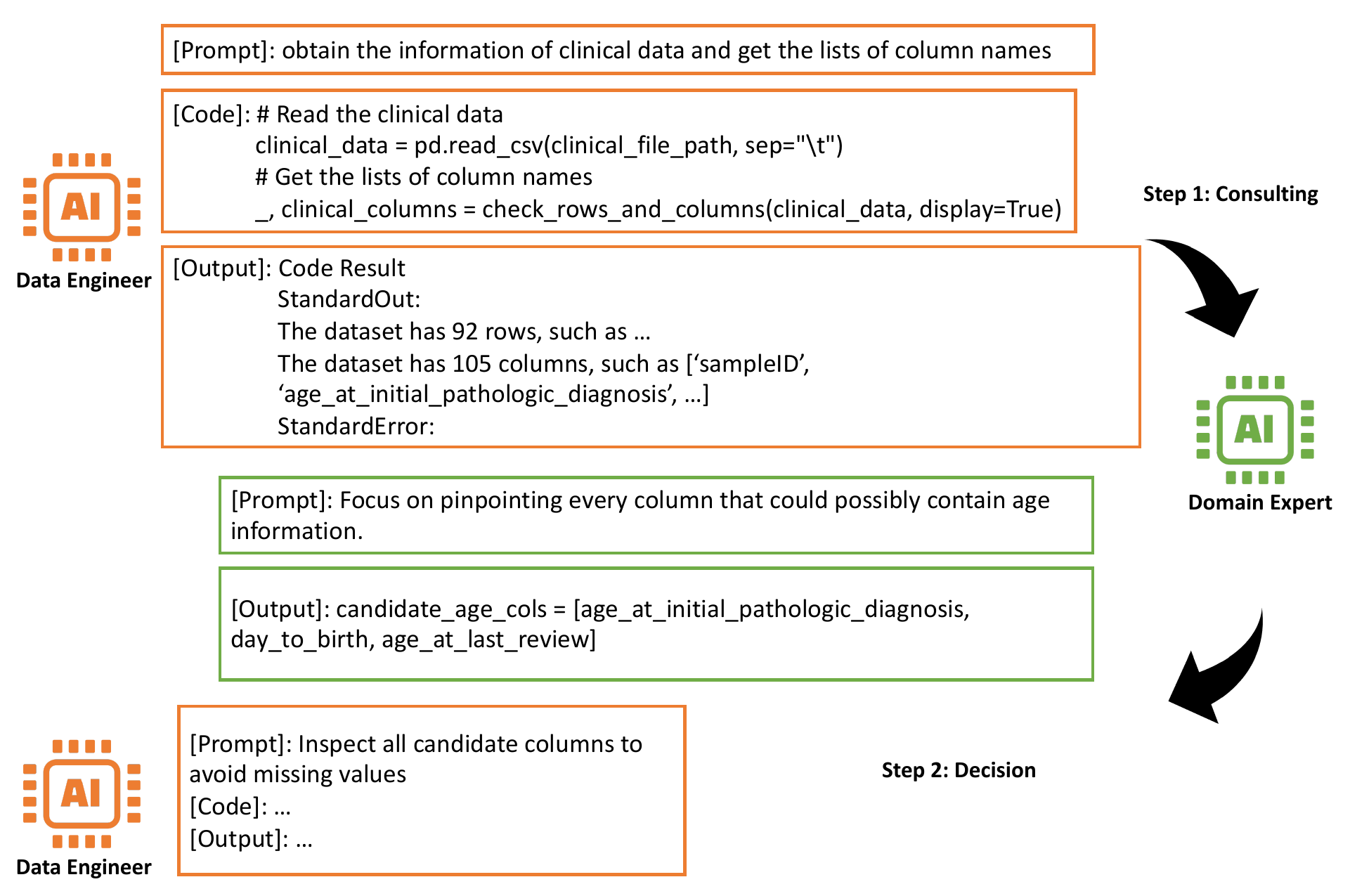}
    \caption{The collaboration between Data Engineer and Domain Expert.}
    \label{fig:domain}
\end{minipage}
\end{figure}

This section discusses the observed limitations of our baseline method, GenoAgent, on the GenoTEX benchmark. We identified that certain steps are inherently challenging, and instability in the feedback mechanism may hinder the agents' iterative improvement process. Figures \ref{fig:reviewer} and \ref{fig:domain} illustrate the two types of feedback mechanisms in GenoAgent.

\subsection{Error example in preprocessing}

The results in Table 4 of the main paper indicate that the preprocessing performance of GenoAgent is primarily constrained by the clinical feature extraction step, which shows a CSC of only 32.28\%. This step is conducted through Domain-Guided Programming (Section 4.2), where the Domain Expert iteratively improves its output based on feedback from the execution environment. Although one round of self-review significantly enhances performance, increasing the maximum review rounds from 1 to 2 yields only marginal benefits. Detailed examination of the agent system’s operation log at this step across different experimental runs reveals that the Domain Expert's answers to the same question can vary randomly. During multiple rounds of self-review, it often provides feedback that contradicts previous suggestions, making it difficult to achieve consistent task performance.

For example, consider the following function used to encode the Breast Cancer trait:
\begin{lstlisting}[language=Python, caption=Failure example of GenoAgent for encoding Breast Cancer trait]
def convert_trait(value):
    if 'breast cancer' in value.lower():
        return 1
    elif 'lung adenocarcinoma' in value.lower():
        return 0
    else:
        return None
\end{lstlisting}

In one run, the code review provided the following feedback:
\begin{lstlisting}[caption=Code review that didn't correctly find the issue\texttt{,} and approved the code]
*Issue*: The convert_trait function assumes that the input string will always mention either "breast cancer" or "lung adenocarcinoma," which might not always be the case. The current data may not explicitly have such a field.

*Suggestion*: Update the function to be more flexible by incorporating biomedical knowledge. Since the dataset deals explicitly with breast cancer and lung adenocarcinoma, we can assume breast cancer is present based on the context or use a default binary value.

*Final Decision*: Approved
\end{lstlisting}

However, in another run with the identical setting, the code review provided different feedback:
\begin{lstlisting}[caption=Another run of code review\texttt{,} which correctly analyzed the issue and rejected the code]
*Issue*: The convert_trait function does not conform to the instructions. The traits should be inferred from the histology field.

*Suggestion*: Adjust the function to check for breast cancer subtypes in the histology field. The current implementation checks for "breast cancer" and "lung adenocarcinoma" in the value, which might not directly map to the provided sample characteristics. A more context-aware approach, utilizing the provided histology information to infer the presence of breast cancer, would be more effective.

*Final Decision*: Rejected
\end{lstlisting}

\subsection{Error example in statistical analysis}
Analysis of failure cases during the statistical analysis task reveals various low-frequency random failures, including errors in extracting data matrices from dataframes and incorrect parameter passing to the regression model. Although no single bottleneck was identified, the cumulative error risk significantly impacts performance, resulting in a suboptimal F$_{1}$ score of 67.08\%. This task involves collaboration between the Statistician and Code Reviewer (Section 4.2). Similar to preprocessing, we observed unstable and inconsistent feedback from the Code Reviewer.

The following is an example of erroneous code generated by the Statistician agent:

\begin{lstlisting}[language=Python, caption=Failure example of the Statistician agent]
from utils.statistics import *

# Set directory and trait, omitted here.

# Step 1: Load input data about the trait into a dataframe
trait_data, _, _ = select_and_load_cohort(data_root, trait, condition=None, is_two_step=False)

# Step 2: Extract X and Y for regression analysis
Y = trait_data[trait].values
X = trait_data.drop(columns=[trait]).values

# Step 3: Selecting appropriate model based on batch effect
batch_effect_detected = detect_batch_effect(X)
model_constructor = LMM if batch_effect_detected else Lasso

# Step 4: Hyperparameter search
param_values = [10 ** i for i in range(-6, 1)]  # powers of 10 from 1e-6 to 1e0
var_names = list(trait_data.columns)
if batch_effect_detected:
    fixed_params = {"fit_intercept": False}
else:
    fixed_params = {"fit_intercept": True, "max_iter": 10000}
best_config, _ = tune_hyperparameters(
    model_constructor, param_values=param_values, X=X, Y=Y, var_names=var_names,
    trait=trait, gene_info_path=gene_info_path, condition=None, Z=None, fixed_params=fixed_params
)

# Step 5: Normalize X
X_normalized, _ = normalize_data(X)

# Step 6: Train the model with best hyperparameter on whole dataset
model = ResidualizationRegressor(model_constructor, best_config)
model.fit(X_normalized, Y)

# Step 7: Interpret trained model for significant genes and save results
significant_genes = interpret_result(model, var_names, trait, condition=None, print_output=True)

os.makedirs(os.path.join(output_dir, normalize_trait(trait)), exist_ok=True)
save_path = os.path.join(output_dir, normalize_trait(trait))
save_result(
    significant_genes, output_dir=save_path, condition=None
)

[Output]:

[Execution Error]:
__init__() got an unexpected keyword argument 'fit_intercept'

\end{lstlisting}

\paragraph{Discussion} The randomness observed may stem from the LLM itself, suggesting a need to prevent one agent from misleading another. During the development of our baseline methods, we implemented several prompt engineering techniques to mitigate this issue: (i) Limiting the Reviewer's feedback to three main suggestions to focus on problem-solving rather than providing numerous distracting comments about code quality, and (ii) Encouraging the agent receiving the review to critically evaluate the feedback and possibly retain its original code. While these measures have alleviated some issues, they persist to some extent in our GenoAgent baseline. A promising future direction involves designing collaborative modes that foster iterative discussions among agents to reconcile differing opinions and enhance their task performance abilities.

We hope this discussion highlights the challenges of our benchmark tasks and encourages future work to address these issues.